\documentclass[10pt,twocolumn,letterpaper]{article}

\usepackage{cvpr}      

\usepackage{multirow}
\usepackage{booktabs}
\usepackage{arydshln} 
\usepackage{graphicx}
\usepackage{pgffor}
\usepackage{tcolorbox}
\usepackage{amsfonts}
\usepackage{inconsolata} 

\definecolor{coralpink}{rgb}{0.97, 0.51, 0.47}
\definecolor{babyblueeyes}{rgb}{0.63, 0.79, 0.95}

\newcommand{\method}{\textsc{CAIRe}\xspace}

\definecolor{cvprblue}{rgb}{0.21,0.49,0.74}
\usepackage[pagebackref,breaklinks,colorlinks,allcolors=cvprblue]{hyperref}

\def\paperID{21700} 
\def\confName{CVPR}
\def\confYear{2026}

\title{\method: Cultural Attribution of Images by Retrieval}

\author{Arnav Yayavaram\thanks{Equal contribution} $^{\,1}$ \; Siddharth Yayavaram$^{*1}$ \; Simran Khanuja$^{*2}$ \; Michael Saxon$^3$ \; Graham Neubig$^2$\\
$^1$BITS Pilani \quad\quad $^2$Carnegie Mellon University \quad\quad $^3$University of California, Santa Barbara \\
{\tt\footnotesize arnav.yayavaram@gmail.com, siddharth.yayavaram@gmail.com, skhanuja@andrew.cmu.edu, michael@saxon.me,  gneubig@cs.cmu.edu}
}

\begin{document}
\maketitle
\begin{abstract}
As text-to-image models become increasingly prevalent, ensuring their equitable performance across diverse cultural contexts is critical. 
Efforts to mitigate cross-cultural biases have been hampered by trade-offs, including a loss in performance, factual inaccuracies, or offensive outputs.
Despite widespread recognition of these challenges,\footnote{For example, in a 2024 incident Google Gemini's T2I diversification techniques produced offensive, historically incorrect images \cite{nyt2024gemini} 
} an inability to reliably measure these biases has stalled progress. To address this gap, we introduce \method\footnote{\url{https://github.com/siddharthyayavaram/CAIRE}}, an evaluation metric that assesses the degree of cultural relevance of an image, given a user-defined set of labels. Our framework grounds entities and concepts in the image to a knowledge base and uses factual information to give independent graded judgments for each culture label.
On a manually curated dataset of culturally salient but rare items built using language models, \method surpasses all baselines by \textbf{22\%} F1 points. Additionally, we construct two datasets for culturally universal concepts, one comprising of T2I generated outputs and another retrieved from naturally-occurring data. \method achieves Pearson’s correlations of \textbf{0.56} and \textbf{0.66} with human ratings on these sets, based on a 5-point Likert scale of cultural relevance. This demonstrates its strong alignment with human judgment across diverse image sources.
\end{abstract}


\begin{figure}[t]
    \centering
    \includegraphics[width=\linewidth]{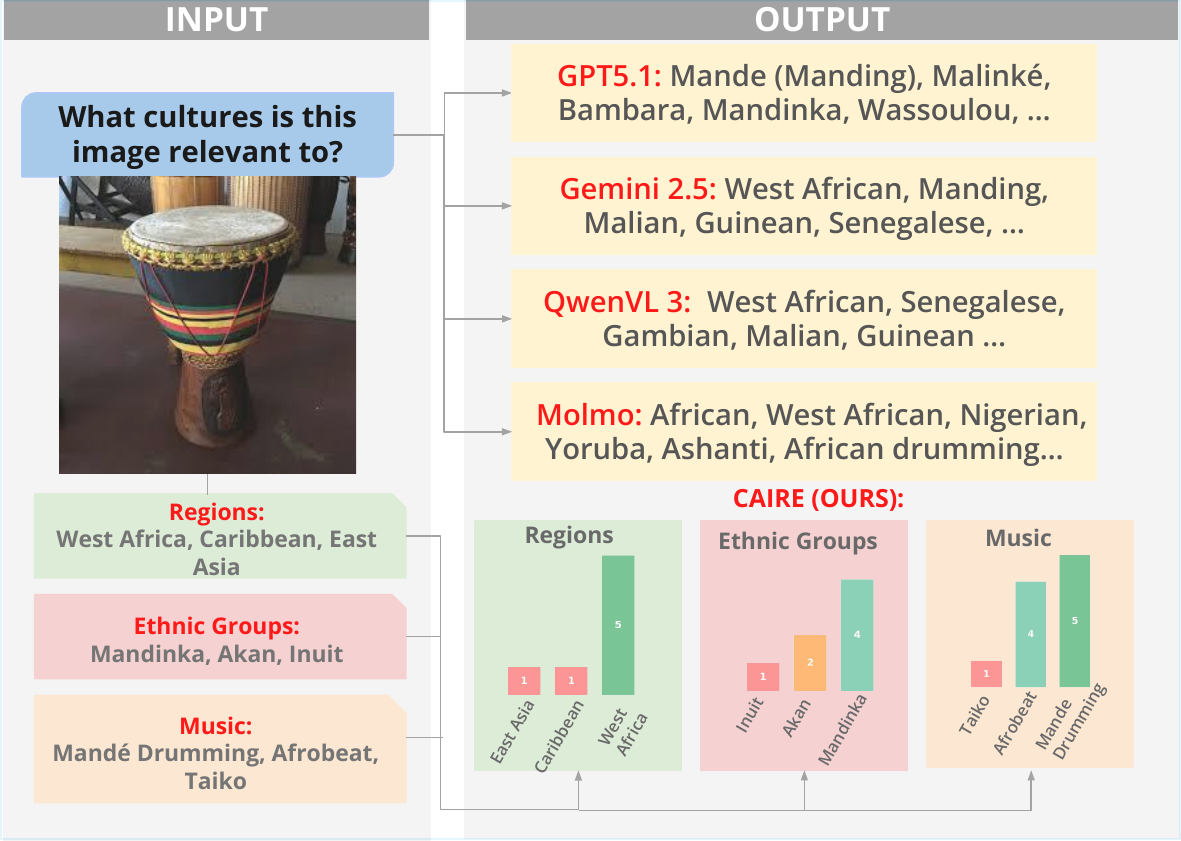}  
    \caption{We propose \method, a novel evaluation metric that assesses the \emph{degree of cultural relevance} of an image, given a user-defined set of labels. Unlike existing methods that assume a definition of culture, we let the user specify cultures to assess as free-text labels (\textit{photo is of a djembe}).
    }
    \label{fig:intro}
\end{figure}

\begin{figure*}[t]
    \centering
    \includegraphics[width=0.95\textwidth]{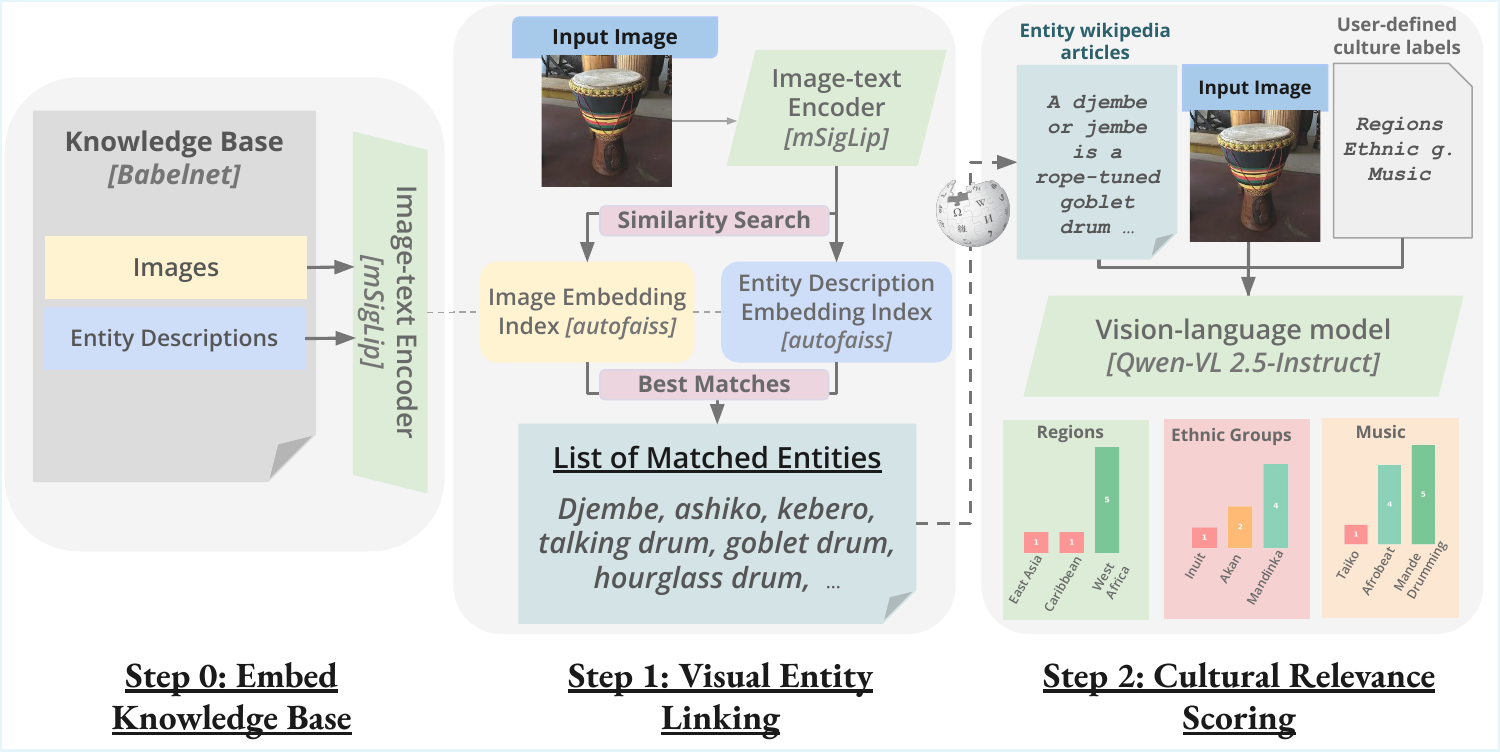}  
    \caption{Overview of \method. From an image-indexed multimodal knowledge base, we embed an input image to retrieve entities that are tied to Wikipedia articles. From the text of those Wikipedia articles and the query image, a vision-language model (VLM) generates an affinity score to each user-specified candidate culture label. A detailed description of our framework is in \S\ref{sec:metric-design}.}
    \label{fig:overview}
\end{figure*}

\section{Introduction}
Current text-to-image (T2I) models, despite their high-quality outputs, have serious issues in cultural representation and sensitivity.
Previous research has reported that they produce culturally homogeneous outputs given under-specified prompts \cite{kannen2024beyond, rassin2024grade} and their outputs are disproportionately biased toward Western cultures, failing to depict global diversity \cite{saxon2023multilingual}.
Further, they generate stereotypical, offensive and factually incorrect outputs when asked to be culturally inclusive \cite{bianchi2023easily,dudley2024opportunity,jha2024visage, wan2024factualitytaxdiversityintervenedtexttoimage}.
The proliferation of AI-generated content, particularly images, is reshaping our digital media ecosystem, with over 15 billion AI-generated images created since 2022.\footnote{ \url{https://journal.everypixel.com/ai-image-statistics} (last accessed: July 29 2025)}

Given this widespread adoption, it is imperative to ensure that these models are inclusive and unbiased towards users from diverse backgrounds.

Moving towards this ideal requires us to first develop robust evaluation metrics that can identify and quantify these cultural representation gaps.
But how do we define \emph{culture}? In the social sciences, culture is a complex concept that can refer to cultural heritage \cite{blake2000defining}, social interactions \cite{monaghan2012cultural}, or ways of life \cite{parsons1971system}. It transcends basic categorizations; spanning ethnicities, communities, and even subtle neighborhood distinctions like those between upper and downtown Manhattan \cite{bucholtz2005identity, eckert2012three}, and is difficult to define concretely because it varies by context. Every individual and group lies at the intersection of multiple cultures (defined by their political, professional, religious, regional, class-based and other affiliations) and these are invoked according to the situation, typically in contrast to other groups \cite{adilazuarda2024towards}. Despite this rich and nuanced complexity, most AI research defaults to using countries as proxies for culture -- indeed, this is one of the most frequently cited limitations across these same papers \cite{zhou2025culture}. Cultural AI work also struggles with capturing culture's inherently dynamic nature. While culture continuously evolves through social negotiation \cite{ochs1996linguistic}, existing benchmarks remain largely static collections of examples or facts \cite{son2024kmmlu,keleg2023dlama,jin2024kobbq,li2023cmmlu}. To address both the fluidity of cultural boundaries and their evolution over time, it is necessary to allow users to define culture on their own terms through natural language descriptions, creating a framework that can adapt as cultures themselves transform.

In this paper we introduce \textbf{C}ultural \textbf{A}ttribution of \textbf{I}mages with \textbf{Re}trieval (\method), a family of evaluation metrics for visual cultural attribution. 
\method evaluates images over user-defined cultural labels, by grounding them to entities and concepts in a knowledge-base (\S\ref{sec:metric-design}).
Our approach takes as input an image and a list of free-text labels representing cultures of interest.
It then outputs a score on a five point scale, indicating the relevance of the image to each label in the culture set
(\autoref{fig:overview}).
The framework operates in two key stages: First, it grounds the image in real-world concepts and objects, leveraging a massively multilingual and multicultural knowledge base.
Second, it utilizes a vision-language model (VLM) to estimate cultural relevance in a retrieval-augmented evaluation setup, where all available information about the recognized concepts is leveraged to estimate this score.
Importantly, our method is designed as a flexible framework, allowing users to integrate their preferred knowledge bases or VLMs to suit their specific needs and contexts.

All previous works which have attempted to evaluate geographical \cite{hall2024digin} or cultural diversity \cite{kannen2024beyond} of images assign a \emph{single} country/region culture label to an image with a binary (relevant / not relevant) score.
In contrast, \method estimates the \emph{degree} of relevance across \emph{multiple} user-defined culture labels.
Due to the absence of test sets with such annotations, we first construct a dataset (\S\ref{sec:test_set}) of rare and culturally significant items, labeled using GPT-4o \cite{openai2024gpt4ocard} and verified with Wikipedia.
On this test set \method surpasses baselines by \textbf{22\%} F1 points.
After validating \method on this dataset, we evaluate its correlation with human judgments of cultural relevance across a broad range of universal concepts. 

\method is adept at capturing annotators' judgments of the relevance of these images to their own cultures, with Pearson's correlations of \textbf{0.56} and \textbf{0.66} to these human ratings over generated and natural images, respectively, demonstrating its alignment with human judgment across diverse image sources (\S\ref{sec:results}). We also demonstrate the use of \method in evaluating cultural diversity across a batch of generated images for a given prompt, highlighting \method's use as a diagnostic tool to reveal T2I models' biases.
\section{Task Formulation}
\label{sec:task}

We define \emph{visual cultural attribution} as the task of assigning cultural relevance scores to an image, given a user-defined set of culture labels. Defining the relevance of an image with respect to a particular culture presents several challenges. An effective framework must:

\begin{enumerate}[noitemsep, left = 0pt]
    \item \textbf{Allow flexible definition of culture labels:} Culture cannot always be represented by simple country or region-based labels; finer-grained community, ethnic, or social group labels must be considered. Thus, it is necessary to allow labels to be defined flexibly using natural language.
    \item \textbf{Provide graded judgments:} Cultural relevance is not binary; many cultural elements are shared across different groups to varying extents.
\end{enumerate}


Formally, given an input image $I$ and a set of cultural labels $\mathcal{C} = \{c_1, c_2, \dots, c_n\}$ (such as the country or religion labels shown in \autoref{fig:overview}), we define the scoring function $f: I \times \mathcal{C} \rightarrow [1, 5]$ that outputs an integer \emph{cultural relevance score} on a five-point scale. A higher score signifies higher cultural relevance of the image to the culture.

Prior work (\S\ref{sec:related}) either assigns binary relevance scores to images or only allows scoring over a fixed cultural proxy label set, thereby failing to adhere to our task formulation.


\section{\method Metric Framework}
\label{sec:metric-design}

In this section, we detail how \method performs visual cultural attribution using information about similar images retrieved from a knowledge base (KB). \method processes each image in two steps: \emph{(a) Visual Entity Linking (VEL)}, which retrieves KB entities relevant to the image using embedding-based matching \cite{Gan2021MultimodalEL}, and \emph{(b) Cultural Relevance Scoring}, which uses VLM judgments over a user-specified set of candidate culture labels. An overview of this pipeline is shown in \autoref{fig:overview} which we detail above.

\subsection{Visual Entity Linking (VEL)}
\label{sec:VEL}

In this step, \method identifies entities and concepts present in the input image by linking them to entries in a knowledge base (KB), which provides rich textual descriptions useful for cultural relevance assessment. Our VEL design follows prior work that leverages vision-language encoders (V-L encoder) for direct image-image and image-text retrieval \cite{sun2022visual, hu2023open}.

To enable retrieval, we first index images and text from a chosen KB using a V-L encoder (see Step 0 in \autoref{fig:overview}). Note that the framework design allows users to integrate custom KBs and V-L encoders, making \method generalizable across domains. For this work, we use the massively multilingual BabelNet graph \cite{navigli2012babelnet} as the KB and mSigLIP  \footnote{\url{https://huggingface.co/google/siglip-so400m-patch16-256-i18n}} \cite{zhai2023sigmoid} as the V-L encoder. BabelNet contains 6 million entities, each associated with images and text descriptions. For each entity, we compute separate mSigLIP embeddings for linked images and text.

These embeddings are stored in two FAISS indices \cite{douze2024faiss}, one for images and one for text, to enable fast similarity search. During inference, given an input image $I$, we compute its embedding $E_I$ and retrieve the top-$K$ most similar KB entities based on cosine similarity:

\begin{equation}
\mathcal{N}(I) = \text{TopK}\big( \text{sim}(E_I, E_{KB}) \big)
\end{equation}

where $\mathcal{N}(I)$ denotes the set of top-$K$ retrieved KB entities, $E_{KB}$ are embeddings of KB images, and $\text{sim}(\cdot, \cdot)$ represents cosine similarity. We empirically choose $K = 20$ in this work as it balances speed and accuracy. Moreover, $K$ is a parameter that can easily be modified in our code.

For clarity, we refer to KB entities as $e$ and candidate culture labels as $c$ throughout the remainder of this paper.

Each retrieved image corresponds to \emph{at least} one KB entity, but may correspond to multiple. For example, a picture of a mango might link to both the \texttt{mango} and \texttt{fruit} nodes in BabelNet---we need to select the most precise and representative entity for cultural assessment. To disambiguate among candidates, we leverage the lemma text associated with each entity. We embed each lemma using the same V-L encoder and compute its similarity to the input image embedding. The final KB entity $e^*$ is selected as:


\begin{equation}
e^* = \arg\max_{e \in \mathcal{N}(I)} \text{sim}(E_I, E_e^{\text{text}})
\end{equation}
where $E_e^{\text{text}}$ is the text embedding of entity $e$ and $\mathcal{N}(I)$ is the set of top-$K$ KB entities retrieved in the previous step.

This two-step retrieval and disambiguation process ensures that the selected KB entity is both visually and semantically aligned with the input image. Empirical evaluation on the FOCI benchmark \cite{geigle2024africaneuropeanswallowbenchmarking} demonstrates the effectiveness of this approach over alternative retrieval-disambiguation techniques, as discussed in Appendix~\S\ref{sec:Vel_appendix}.

\subsection{Cultural Relevance Scoring}
\label{sec:cul_scoring}

Once the input image is linked to relevant KB entities, we assess its cultural relevance using vision-language models (VLMs), conditioned on the candidate culture labels in $\mathcal{C}$. Specifically, we use textual descriptions from Wikipedia articles corresponding to retrieved entities, combined with the input image, to guide the model's judgment.

Rather than asking the model to select a single culture label, we prompt it to independently rate the relevance of the image to each candidate culture in $\mathcal{C}$ on a 1--5 scale. This formulation offers finer granularity and reflects the fact that cultural relevance is rarely binary.

Following prior work on LLMs as judges \citep{li2024llms}, which explore both numerical scoring \cite{jones2024multi} and token-likelihood methods \cite{yuan2021bartscore}, we experimented with two approaches: \textit{(i) numerical scoring} and \textit{(ii) log-likelihood based scoring}. In practice, numerical scoring yielded better alignment with human judgments and proved more straightforward to use. A comparison of both approaches is provided in Appendix~\S\ref{sec:alternative_design}.

In our numerical scoring setup, we prompt the model to produce an integer score between 1 and 5, based on the input Wikipedia text, the image, and a rubric designed to encourage consistent expert-like assessments. An example of the prompt format is shown in Appendix~\S\ref{fig:template_prompt}.

To ensure reliable and well-structured outputs, we apply constrained decoding, restricting the model's output space to valid score tokens (\texttt{1}, \texttt{2}, \texttt{3}, \texttt{4}, \texttt{5}). The final score $s$ is computed as:


\begin{equation}
s = \arg\max_{k \in \{1, 2, 3, 4, 5\}} P(k \mid I, T, c, R)
\end{equation}

Using the model's predicted probability $P(k \mid I, T, c, R)$ of score $k$ given the input image $I$, Wikipedia text $T$, label $c$, and rubric $R$.

This formulation produces interpretable scores that correlate well with human judgments, while remaining efficient and scalable for evaluating cultural relevance across large datasets.

\begin{figure*}[t!]
    \centering

    \begin{subfigure}[b]{0.22\textwidth}
        \centering
        \includegraphics[height=2.5cm]{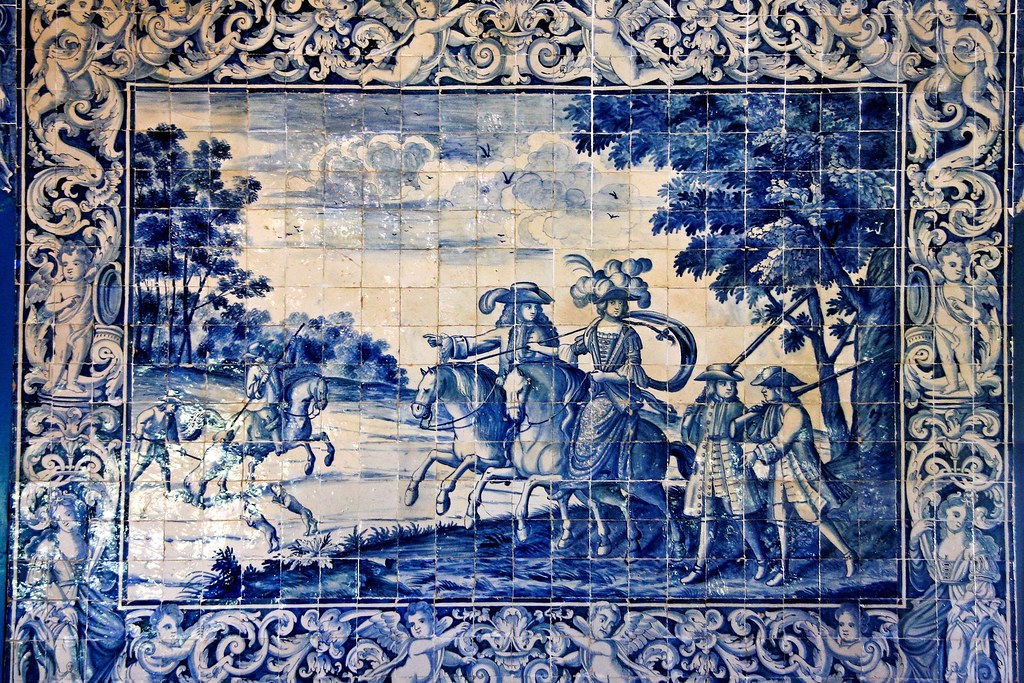}
        \caption{\texttt{\textbf{specific set (countries)}}}
        \label{fig:azulejos}
    \end{subfigure}
    \hfill
    \begin{subfigure}[b]{0.22\textwidth}
        \centering
        \includegraphics[height=2.5cm]{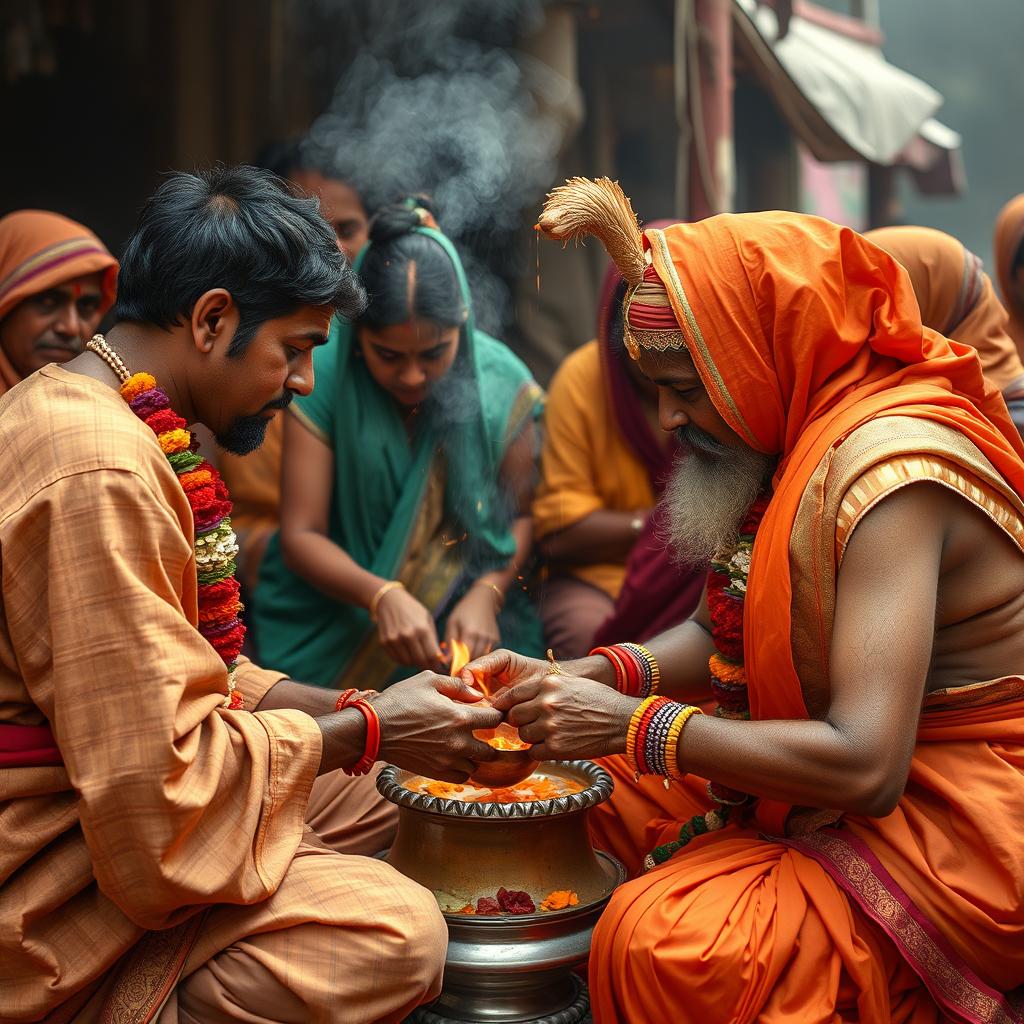}
        \caption{\texttt{\textbf{universal-generated (countries)}}}
        \label{fig:generated}
    \end{subfigure}
    \hfill
    \begin{subfigure}[b]{0.22\textwidth}
        \centering
        \includegraphics[height=2.5cm]{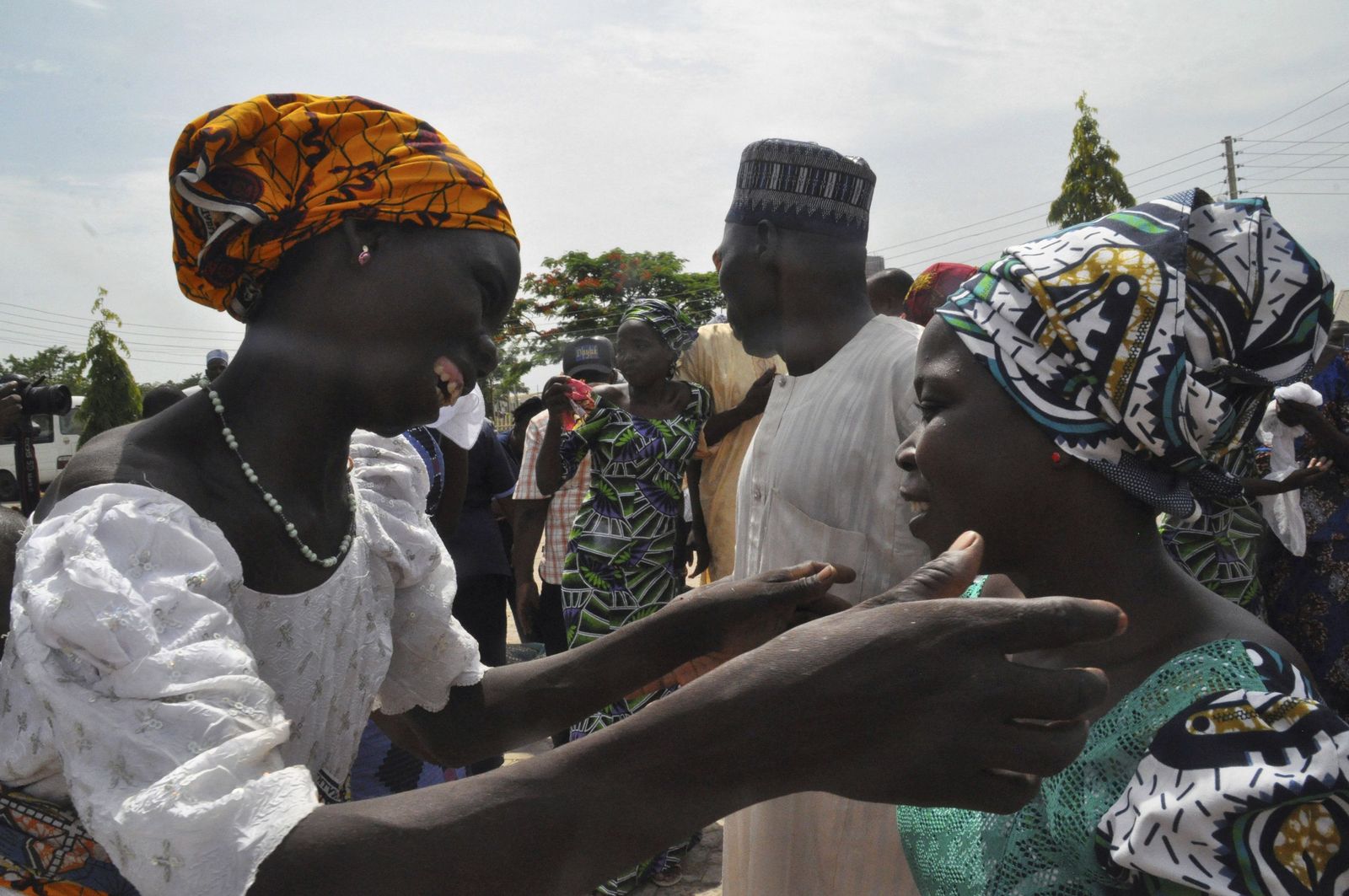}
        \caption{\texttt{\textbf{universal-retrieved (countries)}}}
        \label{fig:retrieved}
    \end{subfigure}

    \medskip

    \begin{subfigure}[b]{0.22\textwidth}
        \centering
        \includegraphics[height=2.5cm]{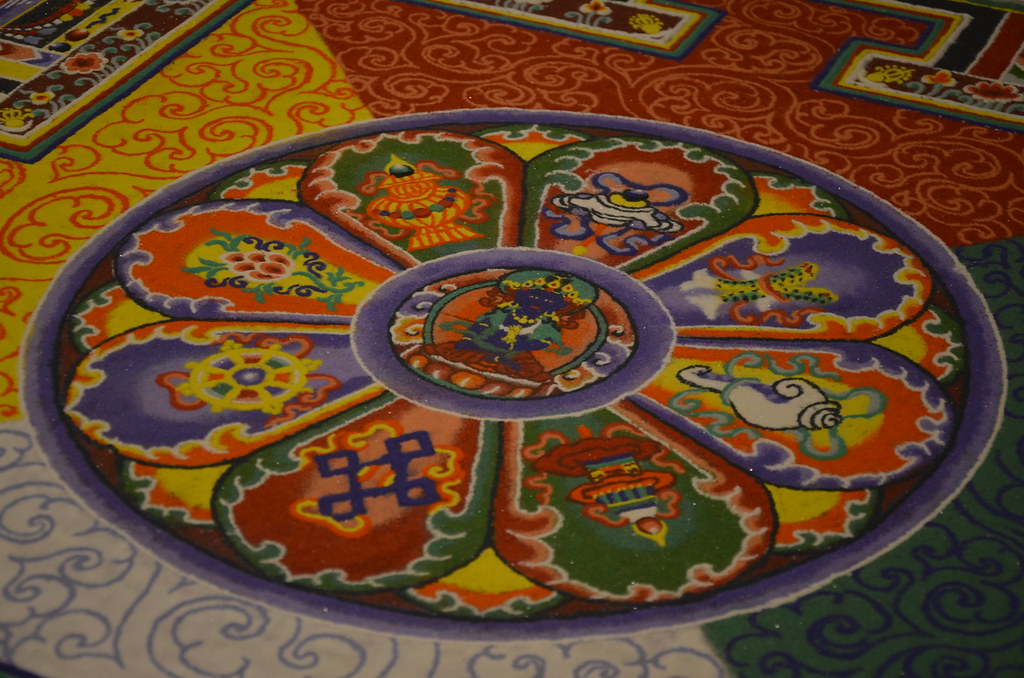}
        \caption{\texttt{\textbf{specific set (religions)}}}
        \label{fig:mandala}
    \end{subfigure}
    \hfill
    \begin{subfigure}[b]{0.22\textwidth}
        \centering
        \includegraphics[height=2.5cm]{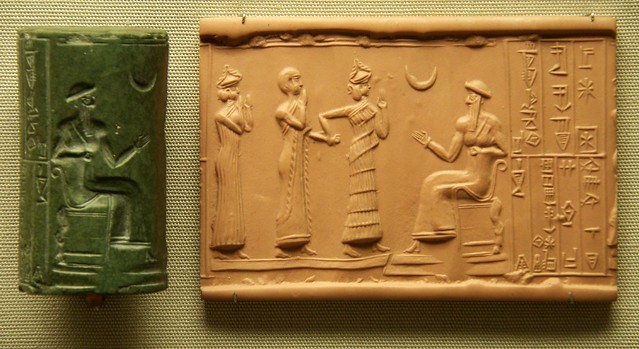}
        \caption{\texttt{\textbf{specific set (Bronze Age civilizations)}}}
        \label{fig:sumercylinder}
    \end{subfigure}
    \hfill
    \begin{subfigure}[b]{0.22\textwidth}
        \centering
        \includegraphics[height=2cm, angle=270]{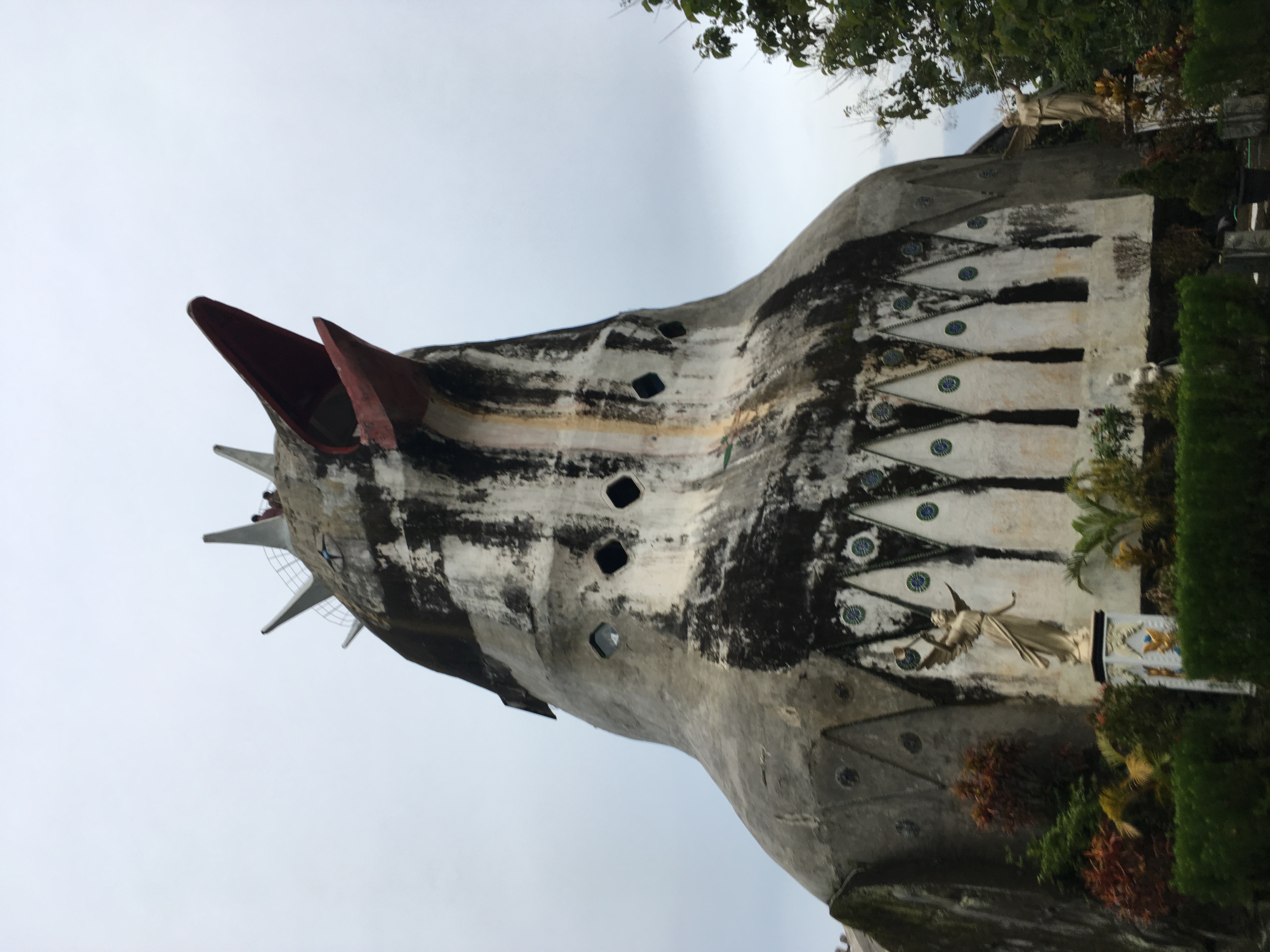}
        \caption{\texttt{\textbf{specific set (cities of Indonesia)}}}
        \label{fig:gereja}
    \end{subfigure}

    \caption{Examples from the evaluation set: (a) image from the \texttt{specific} set depicting Azulejos. The label set of countries consists of \texttt{Portugal}, \texttt{Spain}, \texttt{Brazil}, \texttt{Morocco}, \texttt{Mexico}. (b) T2I-generated image using the prompt ``A realistic photo of a ritual in \texttt{India}," representing the universal-generated subset. (c) image retrieved from DataComp 1B, using the text query ``A realistic photo of greetings in \texttt{Nigeria}", illustrating the universal-retrieved subset. (d-f) additional examples from the \texttt{specific} set corresponding to cultural proxies religion, Bronze Age civilizations, and cities of Indonesia, (Buddhism, Sumer, and Magelang respectively).}
    \label{fig:syn_set_egs}
\end{figure*}

\section{Test Set Curation}
\label{sec:test_set}

Existing datasets used to evaluate cultural and geographic diversity rely solely on single, binary country-level labels \cite{kannen2024beyond, hall2024digin} (\S\ref{sec:related}). In contrast, \S\ref{sec:task} highlights the importance of a dynamic notion of culture—one that allows us to independently measure how relevant each image is to each user-defined cultural category. This renders the label format of current test sets incompatible with our task. To bridge this gap, we introduce two complementary test sets -- \texttt{\textbf{specific}} and \texttt{\textbf{universal}}, each designed to reflect \method’s core desiderata.

The \texttt{specific} set consists of real-world images depicting culturally specific, and often obscure, entities. This set enables us to assess whether \method can reliably capture rare and fine-grained cultural elements. To create labels for this set, we use GPT-4o to assign relevance scores across a diverse set of cultural proxies that extend beyond simple geographic categories, directly supporting \emph{desideratum 1} from \S\ref{sec:task} (examples in Figure \ref{fig:syn_set_egs})

The \texttt{universal} set focuses on culturally universal concepts (e.g., \textit{food}, \textit{ceremony}) and is annotated by 200 annotators from 10 different countries. It includes two splits: \texttt{generated}, containing text-to-image (T2I) generated images, and \texttt{retrieved}, containing natural images sourced to represent each concept. Annotators rated each image on a five-point scale based on its cultural relevance to their own country. Because annotator metadata includes only country-level information, we aggregate these graded judgments across countries. This design satisfies \emph{desideratum 2} from \S\ref{sec:task}, by capturing varying degrees of relevance across diverse cultural perspectives.

\textbf{Why the two-part test set?} The \texttt{specific} set allows us to evaluate \method on highly granular and flexible cultural definitions—such as ethnic groups in Africa or cities in Indonesia—that go beyond coarse national categories. However, these fine-grained labels are difficult to validate with human annotators, as crowdsourcing platforms typically provide only country-level metadata. To address this limitation, we introduce the \texttt{universal} set, which focuses on broad, cross-culturally recognizable concepts (e.g., \textit{breakfast}, \textit{clothing}) for which we can reliably obtain human judgments across many regions. Together, the two sets enable us to assess both \method's ability to understand nuanced cultural definitions and its alignment with human evaluations where such validation is feasible.

\subsection{\texttt{\textbf{specific}} Concept Test Set}




The \texttt{specific} set contains 68 concepts \cite{hurst2024gpt} produced using GPT-4o.
We prompt GPT-4o for rare, culturally distinctive, and visualizable concepts 
according to a range of cultural proxies \cite{adilazuarda2024towards}.
These proxy culture sets include \emph{countries of the world, ethnicities of Africa, states of India, cities of Indonesia, world religions, Native American tribes, and Bronze Age civilizations.}
Within a proxy set, some concepts may be applicable to multiple cultures, for example \textit{shakshuka} to the cuisines of several Maghrebi and Levantine countries.
The complete set is provided in Appendix \S\ref{sec:specific_set_labels}.
Each generated concept and its labels are validated through an affirmative match to an existing Wikipedia article.


We manually collect online creative commons images for each concept, excluding any images present in our KB index to avoid test set contamination. Sample entities are shown in \autoref{fig:syn_set_egs}. 

\paragraph{Metric:} Since GPT-4o does not reliably produce 1--5 numerical ratings, and assigning graded relevance scores is infeasible for many of these rare and culturally specific entities—especially when some cultural groups may be unfamiliar or poorly represented—we instead formulate this as a multi-label classification task. For each image, the goal is to predict the set of culture labels to which the concept is relevant. Performance is measured using F1 score and Precision, computed across all culture labels. This formulation allows us to directly assess whether \method can accurately identify culturally relevant labels for obscure or fine-grained entities. We report both F1 scores and Precision for \method and baseline models in \S\ref{sec:experiments}.

\subsection{\texttt{\textbf{universal}} Concept Test Set}


While the \texttt{specific} test set allows us to test \method's performance on real, rare entities, it does not permit evaluating graded relevance judgments.
Hence, we curate a \texttt{universal} test set, consisting of 20 visually depictable and culturally universal concepts from \cite{bhatia2024local}.
This test set contains both synthetic (\texttt{universal-generated}) and real (\texttt{universal-retrieved}) images of each of these 20 concepts for 10 countries: China, India, the USA, Brazil, Nigeria, Russia, Mexico, Egypt, Germany, and Indonesia.

We generate the synthetic images using Stable Diffusion 3 Medium (SD3) \cite{esser2024scaling} using the prompt \texttt{A realistic photo of [CONCEPT] in [COUNTRY]}.
Using mSIGLIP embeddings we gather images from Datacomp-1B according to this same prompt to acquire the natural images in our test set, which we then manually filter for high-quality photos.

We then hire annotators from these same 10 countries using Prolific\footnote{\url{https://www.prolific.com/}}, and collect cultural relevance ratings for each country, for each image on a five-point Likert scale.

\paragraph{Human Annotation:} We score 400 total images---200 each in
 \texttt{universal-generated} and \texttt{universal-retrieved}. We attempt to hire 20 annotators from all 10 countries to score each image on the 1-5 Likert scale. This way every image receives relevance scores from annotators in every country, ensuring that we not only assess \method's ability to measure relevance, but also \textit{irrelevance}. (Annotator instructions in Appendix \ref{sec:Relevance_Score_appendix_section}, Table \ref{tab:Cultural_relevance_scores}). We provide further details about the human annotation in \S\ref{sec:hum_ann}. 
 




\paragraph{Metric:} We compute Pearson’s correlation coefficient between averaged absolute cultural relevance scores assigned by human raters and those predicted by our tool for each image. 
This metric quantifies the degree of linear association between the predicted scores and human ratings, where a high correlation indicates that our tool not only ranks countries accurately but also assigns scores that align closely with human perceptions of cultural relevance.



\section{Experimental Setup}
\label{sec:experiments}
Since \method is the first work proposing a metric for assessing the degree of cultural relevance of an image, we benchmark its performance against general-purpose V-L encoders and VLMs. We set up these baselines as follows:

\subsection{Baseline 1: Vision-Language Encoders}

The first baseline uses embeddings of the input image and a set of textual \textit{probe prompts} using either CLIP ViT-B/32 \cite{radford2021learning} or mSigLIP \cite{zhai2023sigmoid} to assign a culture label. We include a comparison with alternate V-L Encoders in \S\ref{sec:openclip}.

The probe prompts are constructed from the user-defined set of \textit{candidate culture labels} and a set of five \textit{relevance level} phrases by combining them into the string \texttt{This image is \{relevance level\} to \{culture candidate\}}.
Here, \textit{relevance level} is one of: ``Not relevant", ``Minimally relevant", ``Somewhat relevant", ``Relevant", ``Highly relevant", following the template shown in Table \ref{tab:Cultural_relevance_scores}.

We compute the cosine similarity between the input image and each of the probe prompt embeddings, selecting the label with the highest similarity score as the final rating.


\subsection{Baseline 2: Vision-Language Models} 

We prompt three recent and popular open-source vision-language models (VLMs) -- Llama-3.2-11B-Vision-Instruct \cite{grattafiori2024llama3herdmodels}, Qwen2.5-VL-7B-Instruct \cite{Qwen2VL}, and Pangea-7B-hf \cite{yue2024pangeafullyopenmultilingual} -- as well as one closed-source frontier model, GPT-4o mini, to provide graded scores on cultural relevance.
The prompt includes the target culture along with a description of the relevance levels, as defined in Table \ref{tab:Cultural_relevance_scores}. We encourage the model to follow chain-of-thought reasoning to obtain best results. The full prompt template is provided in  Figure \ref{fig:template_prompt}.

\subsection{Our Metric: \textbf{\method}} 

We test LLaMa-3.2-11B-Vision-Instruct, Qwen2.5-VL-7B-Instruct, and Pangea-7B-hf \cite{grattafiori2024llama3herdmodels, Qwen2VL,yue2024pangeafullyopenmultilingual} as different VLM alternatives in the cultural relevance scoring step of \method (\S\ref{sec:cul_scoring}).


In our experiments, we provide the input image along with the text retrieved during the entity linking (VEL) stage (\S\ref{sec:VEL}) as context to the model.

We experiment with two possible ways of augmenting the VLMs with additional retrieved context:


\begin{itemize}[noitemsep,left = 0pt]
    \item \textit{Wikipedia Text Augmentation:} We incorporate the full Wikipedia page text corresponding to the best matching entity identified through lemma disambiguation. 
    \item \textit{Top-K Entity Title Augmentation:} Instead of detailed descriptions, we provide only the Wikipedia titles of the top 20 retrieved entities. This allows us to test whether models can leverage their own internal knowledge when provided with minimal cues in the form of concept names. This strategy also reflects the fact that retrieval accuracy at the top-20 level is often higher than at the top-1, and thus may offer useful supplementary information.
\end{itemize}

\subsection{Evaluation}

\method and all our baselines, produces a cultural relevance score on a five-point scale, as desired in \S\ref{sec:task}. However, for the \texttt{specific} set, we simply have multi-label binary relevance labels, as explained in \S\ref{sec:specific_set_labels}.
Thus, to evaluate performance on the \texttt{specific} set we convert all metrics' outputs into binary judgments by assigning scores greater than 3 on the five-point scale (\textit{relevant, highly relevant}) to positive label 1, and the others 0.
We report Precision \& F1 scores for each metric based on these in \autoref{tab:results}. We also conduct binary classification at thresholds from 2--5 and observe that \method consistently outperforms all baselines across these thresholds, as detailed in \ref{sec:alt_thresholds}.

For the \texttt{universal} human-labeled test set where we collect ratings on a scale of 1-5, we directly measure Pearson's correlations between the baseline's score and average human ratings, which can be found in \autoref{tab:correlation_countries}.

\section{Results and Analysis}
\label{sec:results}

\begin{table}[t]
    \centering
    \setlength{\tabcolsep}{3.7pt} 
    \resizebox{\columnwidth}{!}{
    \begin{tabular}{l cccc cc}
        \toprule
        \textbf{Model} & \textbf{Img.} & \textbf{Wiki.} & \textbf{Top-20} & \textbf{Prec.} & \textbf{F1} & \textbf{\(\Delta\)$_{\method(F1)}$} \\
        \midrule
        \multicolumn{7}{c}{\textbf{Vision-Language Encoders} (\emph{Baseline-1})} \\
        \midrule
        CLIP & \checkmark &  &  & 5.9 & 3.3 & - \\
        mSigLIP & \checkmark &  &  & 7.7 & 12.8 & - \\
        \midrule
        \multicolumn{7}{c}{\textbf{Vision-Language Models} (\emph{Baseline-2})} \\
        \midrule
        GPT-4o mini & \checkmark  &  &  & 38.7 & 43.1 & - \\
        Llama-3.2-11B-Vis.-Ins. & \checkmark  &  &  & 38.6 & 40.8 & -  \\
        Qwen2.5-VL-7B-Ins. & \checkmark  &  &  & 37.4 & 41.0 & - \\
        Pangea-7B-hf & \checkmark  &  &  & 12.7 & 20.3 & - \\
        \midrule
        \multicolumn{7}{c}{\textbf{\method}} \\
        \midrule
        \addlinespace[3pt] 
        Llama-3.2-11B-Vis.-Ins. & \checkmark  & \checkmark &  & 51.1 & 47.4 & {\scriptsize (+6.6)} \\ 
        & \checkmark  &  & \checkmark & 46.5 & 47.9 & {\scriptsize (+7.1)} \\
        
        
        \hdashline
        \addlinespace[3pt] 
        Qwen2.5-VL-7B-Ins. & \checkmark  & \checkmark &  & \textbf{74.2} & \textbf{65.5} & {\scriptsize (+24.5)} \\ 
        & \checkmark  &  & \checkmark & 64.5 & 54.8 & {\scriptsize (+13.8)} \\
        
        \hdashline
        \addlinespace[3pt] 
        Pangea-7B-hf & \checkmark  & \checkmark &  & 34.3 & 42.9 & {\scriptsize (+22.6)} \\ 
        & \checkmark  &  & \checkmark & 20.6 & 29.0 & {\scriptsize (+8.7)} \\
        \bottomrule
    \end{tabular}
    }
    \caption{
    Precision and F1-scores on the \texttt{specific} set. $\Delta_{\method(F1)}$ represents the improvement in F1-score that \method provides over the naive baseline with that same LM. \textit{Img.} indicates the input image, \textit{Wiki.} represents the Wikipedia content of the top-matched entity, while \textit{Top-20} refers to the names of the top-20 matched entities.
     }
    \label{tab:results}
\end{table}

\begin{table*}[h]
    \centering
    \setlength{\tabcolsep}{5pt} 
    \resizebox{\textwidth}{!}{%
    \begin{tabular}{l c c c c c c c c c c c c c c c c c c c c c c}
        \toprule
        \textbf{Model} & \multicolumn{2}{c}{\textbf{Brazil}} & \multicolumn{2}{c}{\textbf{China}} & \multicolumn{2}{c}{\textbf{Egypt}} & \multicolumn{2}{c}{\textbf{Germany}} & \multicolumn{2}{c}{\textbf{India}} & \multicolumn{2}{c}{\textbf{Indonesia}} & \multicolumn{2}{c}{\textbf{Mexico}} & \multicolumn{2}{c}{\textbf{Nigeria}} & \multicolumn{2}{c}{\textbf{Russia}} & \multicolumn{2}{c}{\textbf{USA}} & \multicolumn{2}{c}{\textbf{Avg.}} \\
        & (N) & (G) & (N) & (G) & (N) & (G) & (N) & (G) & (N) & (G) & (N) & (G) & (N) & (G) & (N) & (G) & (N) & (G) & (N) & (G) & (N) & (G)\\
        \midrule
        \multicolumn{22}{c}{\textbf{Vision-Language Encoders} (\emph{Baseline-1})} \\
        \midrule
        CLIP & -0.04 & 0.38 & 0.04 & 0.16 & 0.09 & 0.07 & -0.17 & -0.09 & -0.05 & 0.12 & 0.28 & 0.33 & -0.06 & 0.07 & -0.02 & 0.20 & -0.10 & 0.08 & -0.02 & 0.11 & -0.02 & 0.11 \\
        mSigLIP & 0.07 & -0.02 & -0.01 & -0.15 & 0.22 & 0.08 & 0.15 & -0.04 & -0.13 & -0.01 & 0.16 & -0.20 & 0.09 & 0.05 & 0.20 & -0.02 & 0.15 & 0.21 & 0.12 & 0.03 & 0.10 & -0.01 \\
        \midrule
        \multicolumn{22}{c}{\textbf{Vision-Language Models} (\emph{Baseline-2})} \\
        \midrule
        
LLaMA3.2-V & 0.54 & 0.18 & 0.57 & 0.37 & 0.31 & 0.29 & 0.65 & 0.27 & 0.59 & 0.34 & 0.56 & 0.33 & 0.45 & 0.11 & 0.68 & 0.56 & 0.57 & 0.01 & 0.52 & 0.28 & 0.55 & 0.31 \\

Qwen2.5-VL & 0.59 & 0.51 & 0.59 & 0.37 & 0.54 & 0.52 & 0.56 & 0.54 & 0.76 & 0.72 & 0.64 & 0.60 & 0.52 & 0.39 & 0.62 & 0.61 & 0.69 & 0.33 & 0.66 & 0.63 & 0.61 & 0.52 \\

Pangea-7B & 0.56 & 0.39 & 0.60 & 0.50 & 0.54 & 0.58 & 0.62 & 0.52 & 0.72 & 0.66 & 0.56 & 0.47 & 0.50 & 0.40 & 0.46 & 0.60 & 0.61 & 0.39 & 0.64 & 0.59 & 0.58 & 0.51 \\
        \midrule
        \multicolumn{22}{c}{\textbf{\method}} \\
        \midrule
        
Llama-3.2-V & 0.64 & 0.47 & 0.61 & \underline{0.30} & 0.52 & 0.49 & 0.70 & 0.58 & 0.69 & 0.70 & 0.65 & 0.44 & 0.57 & 0.33 & 0.47 & 0.49 & \textbf{0.74} & 0.53 & 0.65 & 0.61 & 0.63 & 0.49 \\


Qwen2.5-VL & 0.67 & 0.63 & 0.68 & 0.53 & 0.60 & 0.50 & 0.65 & 0.58 & \textbf{0.74} & 0.68 & 0.70 & 0.53 & 0.58 & \underline{0.46} & 0.61 & 0.55 & \textbf{0.74} & 0.56 & 0.59 & 0.62 & 0.66 & 0.56 \\

Pangea-7B & \textbf{0.70} & 0.49 & 0.64 & 0.45 & 0.61 & 0.56 & 0.62 & 0.51 & 0.52 & 0.65 & 0.58 & 0.55 & 0.59 & \underline{0.39} & 0.64 & 0.64 & 0.61 & 0.47 & 0.61 & 0.47 & 0.61 & 0.52 \\

        \bottomrule
    \end{tabular}}
    \caption{Country-wise Pearson's correlation with human judgment. \textbf{(N)} represents natural data, while \textbf{(G)} represents generated data. The highest and lowest model-wise values are \textbf{bold} and \underline{underlined}, respectively. A detailed analysis of the results is provided in \S\ref{sec:results}. Models used are LLaMA-3.2-11B-Vision-Instruct, Qwen2.5-VL-7B-Instruct, and Pangea-7B-hf.}
    \label{tab:correlation_countries}
\end{table*}

\begin{figure}[t]
    \centering
    
    \begin{subfigure}[t]{0.48\linewidth}
        \centering
        \includegraphics[width=\linewidth]{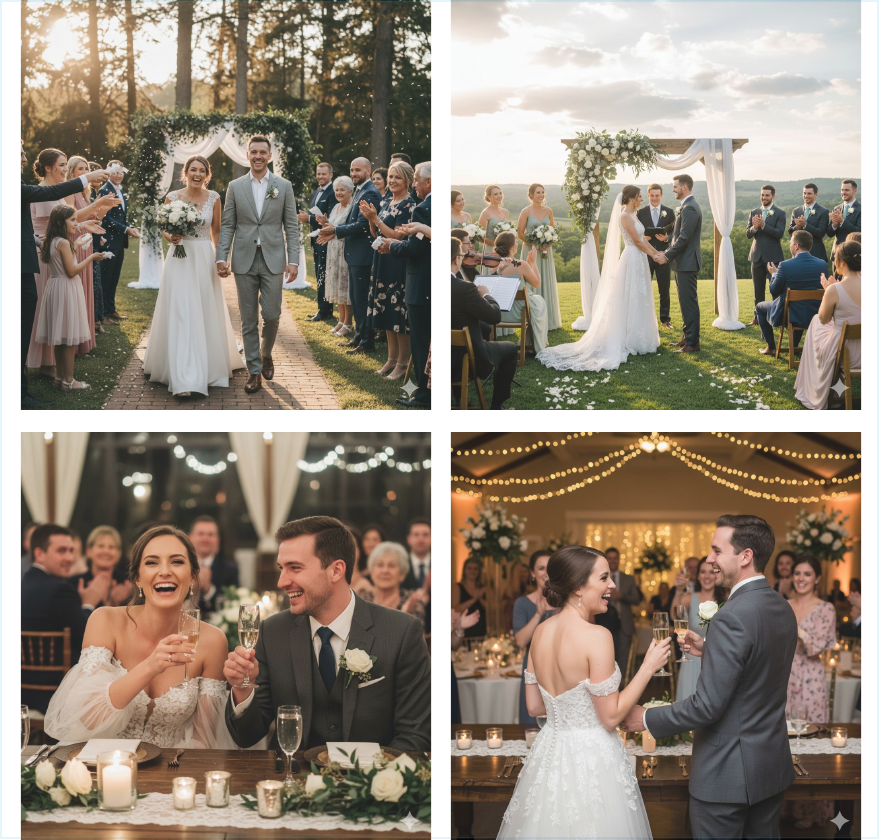}
        \caption{Gemini Nanobanana Outputs}
        \label{fig:sub1}
    \end{subfigure}
    \hfill
    \begin{subfigure}[t]{0.48\linewidth}
        \centering
        \includegraphics[width=\linewidth]{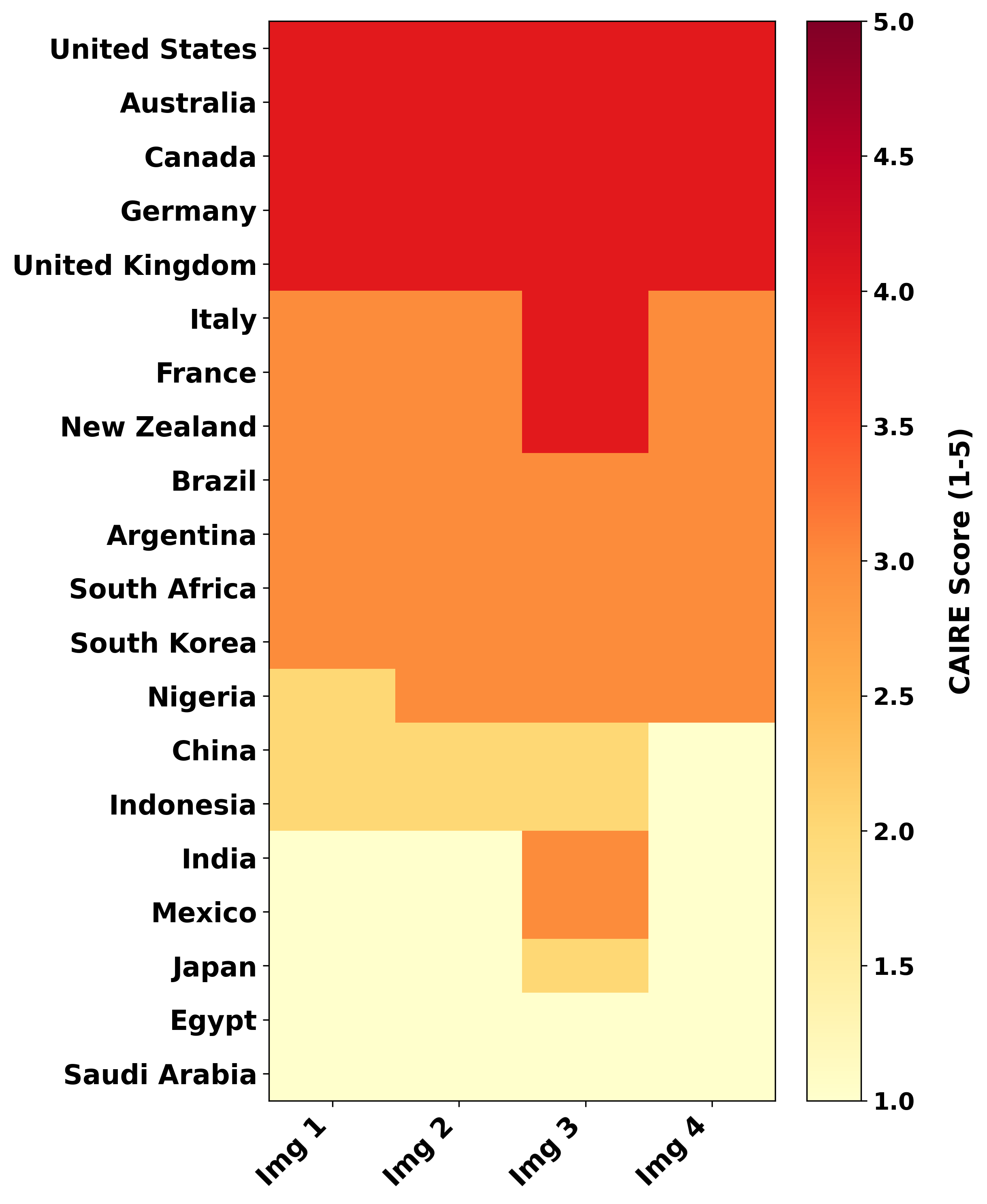}
        \caption{\method score distribution across 20 countries}
        \label{fig:sub2}
    \end{subfigure}
    
    \caption{\method's use in evaluating diversity across T2I generated outputs (prompt: \textit{a photo of a wedding})}
    \label{fig:multiple}
\end{figure}

\autoref{tab:results} presents the Precision \& F1 scores for the \method metrics and baselines on the \texttt{specific} test set. \autoref{tab:correlation_countries} presents the Pearson correlation coefficients of each model against the human-labeled cultural relevance judgments in the \texttt{universal} set.
To analyse results, we ask the following questions:

\paragraph{RQ1: Does \method correctly attribute specific entities to their respective cultures?} 
The \texttt{specific} set checks this using particularly difficult real-world example concepts. 
\autoref{tab:results} shows that \method consistently outperforms both the closed and open-source VLM baselines at this binary cultural relevance classification task. 

Among all open-source VLMs (\textit{Baselines-2}), Qwen2.5-VL-7B-Instruct achieves the highest baseline F1-score of 41.0 without additional context. Integrating this VLM into \method significantly improves performance (by \textbf{25} F1 points!). Although GPT-4o mini remains the strongest standalone baseline with an F1-score of 43.1, incorporating \method into weaker open-source models such as LLaMa-3.2-11B-Vision-Instruct and Qwen2.5-VL-7B-Instruct results in performance that surpasses even this top baseline -- showcasing the effectiveness of \method.



\paragraph{RQ2: How strongly does \method correlate with human opinions?}
The \texttt{universal} set images provide natural subjective annotator opinions of an image's relevance to their own culture. 
Thus, correlation between metric judgments and human labels over this set characterizes how well \method models these opinions.
These results are presented in \autoref{tab:correlation_countries}.
The strongest \method metric on the \texttt{specific} set, Qwen2.5--VL-7B-Instruct, is also most performant on the \texttt{universal} sets, achieving Pearson's correlations of \textbf{0.56} and \textbf{0.66} on average across all countries.
Note that the performance difference between the VLM baselines and \method are not as pronounced on this set---this may be due to the relative commonality of concepts present in the \texttt{universal} images to the \texttt{specific} ones (\autoref{fig:syn_set_egs}).

\paragraph{RQ3: Does \method's model of human opinions generalize across diverse cultures?}
Most columns in \autoref{tab:correlation_countries} contain the correlation scores averaged over all concepts within specific countries.
The lowest and highest values for each model are marked using \underline{underlines} and \textbf{bold} respectively.

 Generally, \method is more performant for India, Germany and Russia compared to Mexico, Egypt and Nigeria. 
This can be attributed to distributional inequalities in mSigLIP's training data and how well-populated the KB is for all these regions. 
This may also be a result of disparities in the representativeness of the presented images in the test set.
We also see that competent VLMs like Qwen display lesser variation across countries (0.5-0.7), while LLaMa displays higher variance. Overall, this suggests that with better models and larger KBs, \method can consistently improve with time.



\paragraph{RQ4: Which variant of \method performs best?}
Qwen2.5-VL-7B-Instruct performs best on the \texttt{specific} set, with models using full Wikipedia pages of top entities outperforming those given only the top 20 entity titles. While multiple titles offer broader context, they can introduce noise, harming fine-grained cultural judgments. On the \texttt{universal} set, Qwen2.5-VL-7B-Instruct again leads, showing the highest average Pearson correlation and lowest cross-country variance across both natural and generated images.

Overall, Qwen models perform strongest with \method, achieving a 25\% absolute F1 gain in performance on the \texttt{specific} set when enhanced with retrieved knowledge.






\paragraph{RQ5: How does performance vary between natural and generated images?} Per \autoref{tab:correlation_countries}, it's clear that \method is more effective in evaluating the cultural relevance of natural images as compared to generated ones. This is expected since arises our KB comprises exclusively natural images and mSigLIP is also trained on naturally-occurring data. Furthermore, the system demonstrates strong performance on the challenging \texttt{specific} set comprising of natural images, reinforcing this observation. In the case of generated images, the depicted entities do not always correspond to real-world concepts or objects, even when the model is explicitly prompted to generate ``realistic" images. However, as generative model capabilities advance, both in terms of realism and in accurately depicting real-world entities, this performance gap is expected to diminish.

\paragraph{RQ6: Can \method help identify cultural skews in batches of generated images?}
Figure \ref{fig:multiple} illustrates how CAIRE scores can be aggregated across a set of generated images to reveal broader cultural patterns in model outputs. Rather than evaluating images individually, examining scores at the batch level highlights which cultures a set of generated samples tends to align with more strongly. In this example, several Western countries (e.g., the United States, United Kingdom, Australia, Germany) consistently receive higher scores across multiple images, suggesting the certain cultural perspectives are systematically better represented than others in the model. \method thus provides a structured way to inspect potential cultural skew in model outputs. 


\section{Related Work}\label{sec:related}

\paragraph{Culturally-diverse image datasets} Several studies have explored the need for culturally diverse image datasets to better assess biases in text-to-image (T2I) models. Recent work \cite{ye2023computer} demonstrated that incorporating culturally and linguistically diverse data in training can enhance the fairness and accuracy of visual representations.
The GeoDE dataset \cite{ramaswamy2023geode} contains 61,940 images of common objects from geographically diverse regions to evaluate object recognition systems otherwise trained on western-centric web-scraped images. \citet{jha2024visage} developed ViSAGe, a dataset of T2I generated outputs, human-annotated for visual stereotypes. \citet{khanuja2024imagespeaksthousandwords} introduce a new task and test set to evaluate whether image-editing models are capable of localizing images for a target culture. Finally, \citet{bhatia2024local} create a benchmark that introduces two challenging tasks to test for cultural inclusion in vision-text models: retrieval across universals and cultural visual grounding. Unlike most of these works that have treated geographical regions as a proxy for culture and assigned single culture labels to each image, our datasets include multi-culture labels with a human rating on how relevant an image might be to each culture in the set. 

\vspace{-10pt}

\paragraph{Evaluating for fairness and diversity} Prior work has attempted to assess biases and fairness in T2I model outputs. DIG In \cite{hall2024digin} and Decomposed-DIG \cite{sureddy2024decomposed} evaluate whether T2I models generate geographically diverse outputs for single-object prompts, revealing representational disparities. \citet{basu2023inspecting} highlighted the Western-centric bias in these models, while \citet{ventura2023navigating} examined embedded cultural perspectives, advocating for improved evaluation frameworks.

CUBE \cite{kannen2024beyond} assesses cultural diversity using a manually curated reference set limited to eight countries and three domains with 300k artifacts. In contrast, our framework leverages a knowledge base spanning 6M concepts, enabling broader assessments. Unlike prior work that primarily evaluates batch-level diversity and focuses only on generated images, our approach assesses cultural relevance at the individual image level and works with both generated and natural images. Finally, while most existing evaluations assign a single country label per image, we introduce a more flexible and inclusive method, allowing multiple cultural proxies to define and assess cultural relevance. Work on multilingualism in text-to-image models overlaps with themes of cultural evaluation. \citet{saxon2023multilingual} introduced an evaluation of cross-lingual capabilities of a slate of T2I models using image similarity over cultural universals.
This line of work has faced challenges for a lack of techniques like \method to perform cultural attribution \cite{saxon2023disparities,saxon2024lost}.

\section{Conclusion}
In this work we introduce \method, a novel and generalizable framework for visual cultural attribution. \method evaluates an image’s relevance to each user-defined culture independently, rather than assigning a single, binary country label to an image as in prior work. \method first uses strong VLMs to identify culturally salient entities represented in the image, grounding it to a large-scale multimodal knowledge base. Using this retrieved information about the image, \method assigns final cultural relevance scores to an image for each culture label, on a scale of 1-5. \method is flexible to a wide range of cultural definitions and applicable across both natural and generated images.

Our experiments show that \method performs strongly in practice: it correlates well with human judgments, outperforms existing baselines across both specific and universal concept sets, and remains robust across diverse cultures. Beyond per-image attribution, we demonstrate that \method functions effectively as a batch-level diversity metric, revealing cultural representation patterns in text-to-image model outputs. Together, these results establish \method as a practical and scalable tool for assessing cultural relevance in vision–language systems.

\section{Limitations \& Ethical Considerations}

\paragraph{Biases from using Wikipedia Content.} Wikipedia is a valuable resource for training language models due to its breadth and structured content, but it also introduces biases. Its coverage reflects the interests of its editor community, leading to variations in topic emphasis and linguistic framing. Additionally, its reliance on verifiable sources can favor well-documented perspectives over emerging viewpoints. These factors influence LM outputs, reinforcing the need for careful interpretation.


\paragraph{Potential Reinforcement of Cultural Stereotypes.} While \method \space aims to provide a nuanced evaluation of cultural relevance, it relies on knowledge bases and vision-language models that may contain biases. If not carefully calibrated, the framework could inadvertently reinforce existing stereotypes by overemphasizing certain cultural elements while under-representing others.

\paragraph{Subjectivity in Cultural Attribution.} Defining cultural relevance is inherently subjective, and different users may have differing perspectives on whether an image aligns with a particular culture.
While \method \space allows for flexible cultural definitions, the subjectivity in user-defined labels and scoring necessitates caution in interpreting results, especially in sensitive contexts.










{
    \small
    \bibliographystyle{ieeenat_fullname}
    \bibliography{main}

\begin{thebibliography}{57}
\providecommand{\natexlab}[1]{#1}
\providecommand{\url}[1]{\texttt{#1}}
\expandafter\ifx\csname urlstyle\endcsname\relax
  \providecommand{\doi}[1]{doi: #1}\else
  \providecommand{\doi}{doi: \begingroup \urlstyle{rm}\Url}\fi

\bibitem[Adilazuarda et~al.(2024)Adilazuarda, Mukherjee, Lavania, Singh, Dwivedi, Aji, O'Neill, Modi, and Choudhury]{adilazuarda2024towards}
Muhammad~Farid Adilazuarda, Sagnik Mukherjee, Pradhyumna Lavania, Siddhant Singh, Ashutosh Dwivedi, Alham~Fikri Aji, Jacki O'Neill, Ashutosh Modi, and Monojit Choudhury.
\newblock Towards measuring and modeling" culture" in llms: A survey.
\newblock \emph{arXiv preprint arXiv:2403.15412}, 2024.

\bibitem[Basu et~al.(2023)Basu, Babu, and Pruthi]{basu2023inspecting}
Abhipsa Basu, R~Venkatesh Babu, and Danish Pruthi.
\newblock Inspecting the geographical representativeness of images from text-to-image models.
\newblock \emph{arXiv preprint arXiv:2305.11080}, 2023.

\bibitem[Bhatia et~al.(2024)Bhatia, Ravi, Chinchure, Hwang, and Shwartz]{bhatia2024local}
Mehar Bhatia, Sahithya Ravi, Aditya Chinchure, Eunjeong Hwang, and Vered Shwartz.
\newblock From local concepts to universals: Evaluating the multicultural understanding of vision-language models.
\newblock \emph{arXiv preprint arXiv:2407.00263}, 2024.

\bibitem[Bianchi et~al.(2023)Bianchi, Kalluri, Durmus, Ladhak, Cheng, Nozza, Hashimoto, Jurafsky, Zou, and Caliskan]{bianchi2023easily}
Federico Bianchi, Pratyusha Kalluri, Esin Durmus, Faisal Ladhak, Myra Cheng, Debora Nozza, Tatsunori Hashimoto, Dan Jurafsky, James Zou, and Aylin Caliskan.
\newblock Easily accessible text-to-image generation amplifies demographic stereotypes at large scale.
\newblock In \emph{Proceedings of the 2023 ACM Conference on Fairness, Accountability, and Transparency}, page 1493–1504, New York, NY, USA, 2023. Association for Computing Machinery.

\bibitem[Blake(2000)]{blake2000defining}
Janet Blake.
\newblock On defining the cultural heritage.
\newblock \emph{International \& Comparative Law Quarterly}, 49\penalty0 (1):\penalty0 61--85, 2000.

\bibitem[Bucholtz and Hall(2005)]{bucholtz2005identity}
Mary Bucholtz and Kira Hall.
\newblock Identity and interaction: A sociocultural linguistic approach.
\newblock \emph{Discourse studies}, 7\penalty0 (4-5):\penalty0 585--614, 2005.

\bibitem[Douze et~al.(2024)Douze, Guzhva, Deng, Johnson, Szilvasy, Mazar{\'e}, Lomeli, Hosseini, and J{\'e}gou]{douze2024faiss}
Matthijs Douze, Alexandr Guzhva, Chengqi Deng, Jeff Johnson, Gergely Szilvasy, Pierre-Emmanuel Mazar{\'e}, Maria Lomeli, Lucas Hosseini, and Herv{\'e} J{\'e}gou.
\newblock The faiss library.
\newblock \emph{arXiv preprint arXiv:2401.08281}, 2024.

\bibitem[Dudley and Kuslikis(2024)]{dudley2024opportunity}
Sean Dudley and Al Kuslikis.
\newblock Opportunity and risk: Artificial intelligence and indian country.
\newblock \emph{Tribal College: Journal of American Indian Higher Education}, 36\penalty0 (2), 2024.

\bibitem[Eckert(2012)]{eckert2012three}
Penelope Eckert.
\newblock Three waves of variation study: The emergence of meaning in the study of sociolinguistic variation.
\newblock \emph{Annual review of Anthropology}, 41\penalty0 (1):\penalty0 87--100, 2012.

\bibitem[Esser et~al.(2024)Esser, Kulal, Blattmann, Entezari, M{\"u}ller, Saini, Levi, Lorenz, Sauer, Boesel, et~al.]{esser2024scaling}
Patrick Esser, Sumith Kulal, Andreas Blattmann, Rahim Entezari, Jonas M{\"u}ller, Harry Saini, Yam Levi, Dominik Lorenz, Axel Sauer, Frederic Boesel, et~al.
\newblock Scaling rectified flow transformers for high-resolution image synthesis.
\newblock In \emph{Forty-first International Conference on Machine Learning}, 2024.

\bibitem[Fang et~al.(2023)Fang, Jose, Jain, Schmidt, Toshev, and Shankar]{fang2023data}
Alex Fang, Albin~Madappally Jose, Amit Jain, Ludwig Schmidt, Alexander Toshev, and Vaishaal Shankar.
\newblock Data filtering networks.
\newblock \emph{arXiv preprint arXiv:2309.17425}, 2023.

\bibitem[Gan et~al.(2021)Gan, Luo, Wang, Wang, He, and Huang]{Gan2021MultimodalEL}
Jingru Gan, Jinchang Luo, Haiwei Wang, Shuhui Wang, W. He, and Qingming Huang.
\newblock Multimodal entity linking: A new dataset and a baseline.
\newblock \emph{Proceedings of the 29th ACM International Conference on Multimedia}, 2021.

\bibitem[Geigle et~al.(2024)Geigle, Timofte, and Glavaš]{geigle2024africaneuropeanswallowbenchmarking}
Gregor Geigle, Radu Timofte, and Goran Glavaš.
\newblock African or european swallow? benchmarking large vision-language models for fine-grained object classification, 2024.

\bibitem[Grant(2024)]{nyt2024gemini}
Nico Grant.
\newblock {Google Chatbot’s A.I. Images Put People of Color in Nazi-Era Uniforms}.
\newblock \emph{{The New York Times}}, 2024.

\bibitem[Grattafiori(2024)]{grattafiori2024llama3herdmodels}
Aaron Grattafiori.
\newblock The llama 3 herd of models, 2024.

\bibitem[Gupta et~al.(2019)Gupta, Dollár, and Girshick]{gupta2019lvisdatasetlargevocabulary}
Agrim Gupta, Piotr Dollár, and Ross Girshick.
\newblock Lvis: A dataset for large vocabulary instance segmentation, 2019.

\bibitem[Hall et~al.(2024)Hall, Ross, Williams, Carion, Drozdzal, and Soriano]{hall2024digin}
Melissa Hall, Candace Ross, Adina Williams, Nicolas Carion, Michal Drozdzal, and Adriana~Romero Soriano.
\newblock Dig in: Evaluating disparities in image generations with indicators for geographic diversity, 2024.

\bibitem[Hu et~al.(2023)Hu, Luan, Chen, Khandelwal, Joshi, Lee, Toutanova, and Chang]{hu2023open}
Hexiang Hu, Yi Luan, Yang Chen, Urvashi Khandelwal, Mandar Joshi, Kenton Lee, Kristina Toutanova, and Ming-Wei Chang.
\newblock Open-domain visual entity recognition: Towards recognizing millions of wikipedia entities.
\newblock In \emph{Proceedings of the IEEE/CVF International Conference on Computer Vision}, pages 12065--12075, 2023.

\bibitem[Hurst et~al.(2024)Hurst, Lerer, Goucher, Perelman, Ramesh, Clark, Ostrow, Welihinda, Hayes, Radford, et~al.]{hurst2024gpt}
Aaron Hurst, Adam Lerer, Adam~P Goucher, Adam Perelman, Aditya Ramesh, Aidan Clark, AJ Ostrow, Akila Welihinda, Alan Hayes, Alec Radford, et~al.
\newblock Gpt-4o system card.
\newblock \emph{arXiv preprint arXiv:2410.21276}, 2024.

\bibitem[Jha et~al.(2024)Jha, Prabhakaran, Denton, Laszlo, Dave, Qadri, Reddy, and Dev]{jha2024visage}
Akshita Jha, Vinodkumar Prabhakaran, Remi Denton, Sarah Laszlo, Shachi Dave, Rida Qadri, Chandan~K. Reddy, and Sunipa Dev.
\newblock Visage: A global-scale analysis of visual stereotypes in text-to-image generation, 2024.

\bibitem[Jin et~al.(2024)Jin, Kim, Lee, Yoo, Oh, and Lee]{jin2024kobbq}
Jiho Jin, Jiseon Kim, Nayeon Lee, Haneul Yoo, Alice Oh, and Hwaran Lee.
\newblock Kobbq: Korean bias benchmark for question answering.
\newblock \emph{Transactions of the Association for Computational Linguistics}, 12:\penalty0 507--524, 2024.

\bibitem[Jones et~al.(2024)Jones, Mo, Fosler-Lussier, and Sun]{jones2024multi}
Jaylen Jones, Lingbo Mo, Eric Fosler-Lussier, and Huan Sun.
\newblock A multi-aspect framework for counter narrative evaluation using large language models.
\newblock \emph{arXiv preprint arXiv:2402.11676}, 2024.

\bibitem[Kannen et~al.(2024)Kannen, Ahmad, Andreetto, Prabhakaran, Prabhu, Dieng, Bhattacharyya, and Dave]{kannen2024beyond}
Nithish Kannen, Arif Ahmad, Marco Andreetto, Vinodkumar Prabhakaran, Utsav Prabhu, Adji~Bousso Dieng, Pushpak Bhattacharyya, and Shachi Dave.
\newblock Beyond aesthetics: Cultural competence in text-to-image models.
\newblock \emph{arXiv preprint arXiv:2407.06863}, 2024.

\bibitem[Keleg and Magdy(2023)]{keleg2023dlama}
Amr Keleg and Walid Magdy.
\newblock Dlama: A framework for curating culturally diverse facts for probing the knowledge of pretrained language models.
\newblock \emph{arXiv preprint arXiv:2306.05076}, 2023.

\bibitem[Khanuja et~al.(2024)Khanuja, Ramamoorthy, Song, and Neubig]{khanuja2024imagespeaksthousandwords}
Simran Khanuja, Sathyanarayanan Ramamoorthy, Yueqi Song, and Graham Neubig.
\newblock An image speaks a thousand words, but can everyone listen? on image transcreation for cultural relevance, 2024.

\bibitem[Laurençon et~al.(2024)Laurençon, Tronchon, Cord, and Sanh]{laurençon2024matters}
Hugo Laurençon, Léo Tronchon, Matthieu Cord, and Victor Sanh.
\newblock What matters when building vision-language models?, 2024.

\bibitem[Li et~al.(2023)Li, Zhang, Koto, Yang, Zhao, Gong, Duan, and Baldwin]{li2023cmmlu}
Haonan Li, Yixuan Zhang, Fajri Koto, Yifei Yang, Hai Zhao, Yeyun Gong, Nan Duan, and Timothy Baldwin.
\newblock Cmmlu: Measuring massive multitask language understanding in chinese.
\newblock \emph{arXiv preprint arXiv:2306.09212}, 2023.

\bibitem[Li et~al.(2024)Li, Dong, Chen, Su, Zhou, Ai, Ye, and Liu]{li2024llms}
Haitao Li, Qian Dong, Junjie Chen, Huixue Su, Yujia Zhou, Qingyao Ai, Ziyi Ye, and Yiqun Liu.
\newblock Llms-as-judges: a comprehensive survey on llm-based evaluation methods.
\newblock \emph{arXiv preprint arXiv:2412.05579}, 2024.

\bibitem[Lin et~al.(2015)Lin, Maire, Belongie, Bourdev, Girshick, Hays, Perona, Ramanan, Zitnick, and Dollár]{lin2015microsoftcococommonobjects}
Tsung-Yi Lin, Michael Maire, Serge Belongie, Lubomir Bourdev, Ross Girshick, James Hays, Pietro Perona, Deva Ramanan, C.~Lawrence Zitnick, and Piotr Dollár.
\newblock Microsoft coco: Common objects in context, 2015.

\bibitem[Liu et~al.(2024)Liu, Li, Li, Li, Zhang, Shen, and Lee]{liu2024llavanext}
Haotian Liu, Chunyuan Li, Yuheng Li, Bo Li, Yuanhan Zhang, Sheng Shen, and Yong~Jae Lee.
\newblock Llava-next: Improved reasoning, ocr, and world knowledge, 2024.

\bibitem[Monaghan et~al.(2012)Monaghan, Goodman, and Robinson]{monaghan2012cultural}
Leila Monaghan, Jane~E Goodman, and Jennifer Robinson.
\newblock \emph{A cultural approach to interpersonal communication: Essential readings}.
\newblock John Wiley \& Sons, 2012.

\bibitem[Navigli and Ponzetto(2012)]{navigli2012babelnet}
Roberto Navigli and Simone~Paolo Ponzetto.
\newblock Babelnet: The automatic construction, evaluation and application of a wide-coverage multilingual semantic network.
\newblock \emph{Artificial intelligence}, 193:\penalty0 217--250, 2012.

\bibitem[Ochs(1996)]{ochs1996linguistic}
Elinor Ochs.
\newblock \emph{Linguistic resources for socializing humanity.}
\newblock Cambridge University Press, 1996.

\bibitem[OpenAI(2024)]{openai2024gpt4ocard}
Aaron~Hurst OpenAI.
\newblock Gpt-4o system card, 2024.

\bibitem[Oquab et~al.(2024)Oquab, Darcet, Moutakanni, Vo, Szafraniec, Khalidov, Fernandez, Haziza, Massa, El-Nouby, Assran, Ballas, Galuba, Howes, Huang, Li, Misra, Rabbat, Sharma, Synnaeve, Xu, Jegou, Mairal, Labatut, Joulin, and Bojanowski]{oquab2024dinov2learningrobustvisual}
Maxime Oquab, Timothée Darcet, Théo Moutakanni, Huy Vo, Marc Szafraniec, Vasil Khalidov, Pierre Fernandez, Daniel Haziza, Francisco Massa, Alaaeldin El-Nouby, Mahmoud Assran, Nicolas Ballas, Wojciech Galuba, Russell Howes, Po-Yao Huang, Shang-Wen Li, Ishan Misra, Michael Rabbat, Vasu Sharma, Gabriel Synnaeve, Hu Xu, Hervé Jegou, Julien Mairal, Patrick Labatut, Armand Joulin, and Piotr Bojanowski.
\newblock Dinov2: Learning robust visual features without supervision, 2024.

\bibitem[Parsons(1971)]{parsons1971system}
Talcott Parsons.
\newblock The system of modern societies, 1971.

\bibitem[Qi et~al.(2023)Qi, Xie, Li, Ge, and Zhang]{qi2023balanced}
Tianhao Qi, Hongtao Xie, Pandeng Li, Jiannan Ge, and Yongdong Zhang.
\newblock Balanced classification: A unified framework for long-tailed object detection.
\newblock \emph{IEEE Transactions on Multimedia}, 26:\penalty0 3088--3101, 2023.

\bibitem[Radford et~al.(2021)Radford, Kim, Hallacy, Ramesh, Goh, Agarwal, Sastry, Askell, Mishkin, Clark, et~al.]{radford2021learning}
Alec Radford, Jong~Wook Kim, Chris Hallacy, Aditya Ramesh, Gabriel Goh, Sandhini Agarwal, Girish Sastry, Amanda Askell, Pamela Mishkin, Jack Clark, et~al.
\newblock Learning transferable visual models from natural language supervision.
\newblock In \emph{International conference on machine learning}, pages 8748--8763. PMLR, 2021.

\bibitem[Ramaswamy et~al.(2023)Ramaswamy, Lin, Zhao, Adcock, van~der Maaten, Ghadiyaram, and Russakovsky]{ramaswamy2023geode}
Vikram~V Ramaswamy, Sing~Yu Lin, Dora Zhao, Aaron Adcock, Laurens van~der Maaten, Deepti Ghadiyaram, and Olga Russakovsky.
\newblock Geode: a geographically diverse evaluation dataset for object recognition.
\newblock \emph{Advances in Neural Information Processing Systems}, 36:\penalty0 66127--66137, 2023.

\bibitem[Rassin et~al.(2024)Rassin, Slobodkin, Ravfogel, Elazar, and Goldberg]{rassin2024grade}
Royi Rassin, Aviv Slobodkin, Shauli Ravfogel, Yanai Elazar, and Yoav Goldberg.
\newblock Grade: Quantifying sample diversity in text-to-image models.
\newblock \emph{arXiv preprint arXiv:2410.22592}, 2024.

\bibitem[Saxon and Wang(2023{\natexlab{a}})]{saxon2023disparities}
Michael Saxon and William~Yang Wang.
\newblock Disparities in text-to-image model concept possession across languages.
\newblock In \emph{Proceedings of the 2023 ACM Conference on Fairness, Accountability, and Transparency}, page 1870, New York, NY, USA, 2023{\natexlab{a}}. Association for Computing Machinery.

\bibitem[Saxon and Wang(2023{\natexlab{b}})]{saxon2023multilingual}
Michael Saxon and William~Yang Wang.
\newblock Multilingual conceptual coverage in text-to-image models.
\newblock In \emph{Proceedings of the 61st Annual Meeting of the Association for Computational Linguistics (Volume 1: Long Papers)}, pages 4831--4848, Toronto, Canada, 2023{\natexlab{b}}. Association for Computational Linguistics.

\bibitem[Saxon et~al.(2024)Saxon, Luo, Levy, Baral, Yang, and Wang]{saxon2024lost}
Michael Saxon, Yiran Luo, Sharon Levy, Chitta Baral, Yezhou Yang, and William~Yang Wang.
\newblock Lost in translation? translation errors and challenges for fair assessment of text-to-image models on multilingual concepts.
\newblock In \emph{Proceedings of the 2024 Conference of the North American Chapter of the Association for Computational Linguistics: Human Language Technologies (Volume 2: Short Papers)}, pages 572--582, Mexico City, Mexico, 2024. Association for Computational Linguistics.

\bibitem[Son et~al.(2024)Son, Lee, Kim, Kim, Muennighoff, Choi, Park, Yoo, and Biderman]{son2024kmmlu}
Guijin Son, Hanwool Lee, Sungdong Kim, Seungone Kim, Niklas Muennighoff, Taekyoon Choi, Cheonbok Park, Kang~Min Yoo, and Stella Biderman.
\newblock Kmmlu: Measuring massive multitask language understanding in korean.
\newblock \emph{arXiv preprint arXiv:2402.11548}, 2024.

\bibitem[Sun et~al.(2022)Sun, Fan, Guo, Zhang, and Cheng]{sun2022visual}
Wenxiang Sun, Yixing Fan, Jiafeng Guo, Ruqing Zhang, and Xueqi Cheng.
\newblock Visual named entity linking: A new dataset and a baseline.
\newblock \emph{arXiv preprint arXiv:2211.04872}, 2022.

\bibitem[Sureddy et~al.(2024)Sureddy, Padalia, Periyakaruppa, Saha, Williams, Romero-Soriano, Richards, Kirichenko, and Hall]{sureddy2024decomposed}
Abhishek Sureddy, Dishant Padalia, Nandhinee Periyakaruppa, Oindrila Saha, Adina Williams, Adriana Romero-Soriano, Megan Richards, Polina Kirichenko, and Melissa Hall.
\newblock Decomposed evaluations of geographic disparities in text-to-image models.
\newblock \emph{arXiv preprint arXiv:2406.11988}, 2024.

\bibitem[Ventura et~al.(2023)Ventura, Ben-David, Korhonen, and Reichart]{ventura2023navigating}
Mor Ventura, Eyal Ben-David, Anna Korhonen, and Roi Reichart.
\newblock Navigating cultural chasms: Exploring and unlocking the cultural pov of text-to-image models.
\newblock \emph{arXiv preprint arXiv:2310.01929}, 2023.

\bibitem[Wan et~al.(2024)Wan, Wu, Wang, and Chang]{wan2024factualitytaxdiversityintervenedtexttoimage}
Yixin Wan, Di Wu, Haoran Wang, and Kai-Wei Chang.
\newblock The factuality tax of diversity-intervened text-to-image generation: Benchmark and fact-augmented intervention, 2024.

\bibitem[Wang et~al.(2024{\natexlab{a}})Wang, Chen, Liu, Chen, Lin, Han, and Ding]{wang2024yolov10realtimeendtoendobject}
Ao Wang, Hui Chen, Lihao Liu, Kai Chen, Zijia Lin, Jungong Han, and Guiguang Ding.
\newblock Yolov10: Real-time end-to-end object detection, 2024{\natexlab{a}}.

\bibitem[Wang et~al.(2024{\natexlab{b}})Wang, Bai, Tan, Wang, Fan, Bai, Chen, Liu, Wang, Ge, Fan, Dang, Du, Ren, Men, Liu, Zhou, Zhou, and Lin]{Qwen2VL}
Peng Wang, Shuai Bai, Sinan Tan, Shijie Wang, Zhihao Fan, Jinze Bai, Keqin Chen, Xuejing Liu, Jialin Wang, Wenbin Ge, Yang Fan, Kai Dang, Mengfei Du, Xuancheng Ren, Rui Men, Dayiheng Liu, Chang Zhou, Jingren Zhou, and Junyang Lin.
\newblock Qwen2-vl: Enhancing vision-language model's perception of the world at any resolution.
\newblock \emph{arXiv preprint arXiv:2409.12191}, 2024{\natexlab{b}}.

\bibitem[Wu et~al.(2019)Wu, Kirillov, Massa, Lo, and Girshick]{wu2019detectron2}
Yuxin Wu, Alexander Kirillov, Francisco Massa, Wan-Yen Lo, and Ross Girshick.
\newblock Detectron2.
\newblock \url{https://github.com/facebookresearch/detectron2}, 2019.

\bibitem[Ye et~al.(2023)Ye, Santy, Hwang, Zhang, and Krishna]{ye2023computer}
Andre Ye, Sebastin Santy, Jena~D Hwang, Amy~X Zhang, and Ranjay Krishna.
\newblock Computer vision datasets and models exhibit cultural and linguistic diversity in perception.
\newblock \emph{arXiv preprint arXiv}, 2310:\penalty0 4, 2023.

\bibitem[Yuan et~al.(2021)Yuan, Neubig, and Liu]{yuan2021bartscore}
Weizhe Yuan, Graham Neubig, and Pengfei Liu.
\newblock Bartscore: evaluating generated text as text generation.
\newblock In \emph{Proceedings of the 35th International Conference on Neural Information Processing Systems}, Red Hook, NY, USA, 2021. Curran Associates Inc.

\bibitem[Yue et~al.(2024)Yue, Song, Asai, Kim, de~Dieu~Nyandwi, Khanuja, Kantharuban, Sutawika, Ramamoorthy, and Neubig]{yue2024pangeafullyopenmultilingual}
Xiang Yue, Yueqi Song, Akari Asai, Seungone Kim, Jean de Dieu~Nyandwi, Simran Khanuja, Anjali Kantharuban, Lintang Sutawika, Sathyanarayanan Ramamoorthy, and Graham Neubig.
\newblock Pangea: A fully open multilingual multimodal llm for 39 languages.
\newblock \emph{arXiv preprint arXiv:2410.16153}, 2024.

\bibitem[Zhai et~al.(2023)Zhai, Mustafa, Kolesnikov, and Beyer]{zhai2023sigmoid}
Xiaohua Zhai, Basil Mustafa, Alexander Kolesnikov, and Lucas Beyer.
\newblock Sigmoid loss for language image pre-training, 2023.

\bibitem[Zhao et~al.(2021)Zhao, Wallace, Feng, Klein, and Singh]{zhao2021calibrate}
Zihao Zhao, Eric Wallace, Shi Feng, Dan Klein, and Sameer Singh.
\newblock Calibrate before use: Improving few-shot performance of language models.
\newblock In \emph{International conference on machine learning}, pages 12697--12706. PMLR, 2021.

\bibitem[Zhou et~al.(2025)Zhou, Bamman, and Bleaman]{zhou2025culture}
Naitian Zhou, David Bamman, and Isaac~L Bleaman.
\newblock Culture is not trivia: Sociocultural theory for cultural nlp.
\newblock \emph{arXiv preprint arXiv:2502.12057}, 2025.

\end{thebibliography}
}


\clearpage
\appendix

\section{Details on Visual Entity Linking}

\subsection{Off-the-shelf Object Detection}

In our approach to Visual Entity Linking (VEL), we initially employed a straightforward methodology by leveraging off-the-shelf object detection models to identify entities within images. We utilized Detectron 2 \cite{wu2019detectron2} from Facebook research along with YOLOv10 \cite{wang2024yolov10realtimeendtoendobject}. The underlying premise was to recognize object names through these detection tools and subsequently map them to corresponding knowledge base (KB) identifiers, akin to traditional textual entity linking. To this end, we experimented with state-of-the-art object detection models capable of recognizing fine-grained object categories in both open and closed-world settings, where KB identifiers served as predefined classification labels. COCO \cite{lin2015microsoftcococommonobjects} and LVIS \cite{gupta2019lvisdatasetlargevocabulary} classes are 2 of the most commonly used sets of classes. Additionally, we utilized Vision language models to detect entities. Specifically, we used LlaVA-NeXT \cite{liu2024llavanext} and Idefics2-8B \cite{laurençon2024matters}.

However, these methods yielded suboptimal performance, particularly for long-tail entities, a challenge commonly observed in object detection \cite{qi2023balanced}. Given that culturally rare or niche objects often fall within the long-tail distribution, this approach was deemed impractical for robust entity linking.\footnote{A comparative analysis of the outputs produced by various object detection models and VLMs utilized in our study can be found here:  
\href{https://hub.zenoml.com/project/954d212c-eca6-4ad6-a461-7cb623641732/Object_Captioning}{https://hub.zenoml.com/project/954d212c-eca6-4ad6-a461-7cb623641732/Object Captioning}. We present the results for 5 different systems; Using Detectron2, yolo10x, LlaVA-NeXT, Idefics2-8B, and an Entity-Attribute-Relation (EAR) pipeline.} The results are presented on the \textit{``transcreation"} dataset introduced by \cite{khanuja2024imagespeaksthousandwords}.

\subsection{Encoder Models for Building KB Index}

\begin{figure}[htbp]
\centering
\begin{subfigure}[b]{0.22\textwidth}
\includegraphics[width=\textwidth]{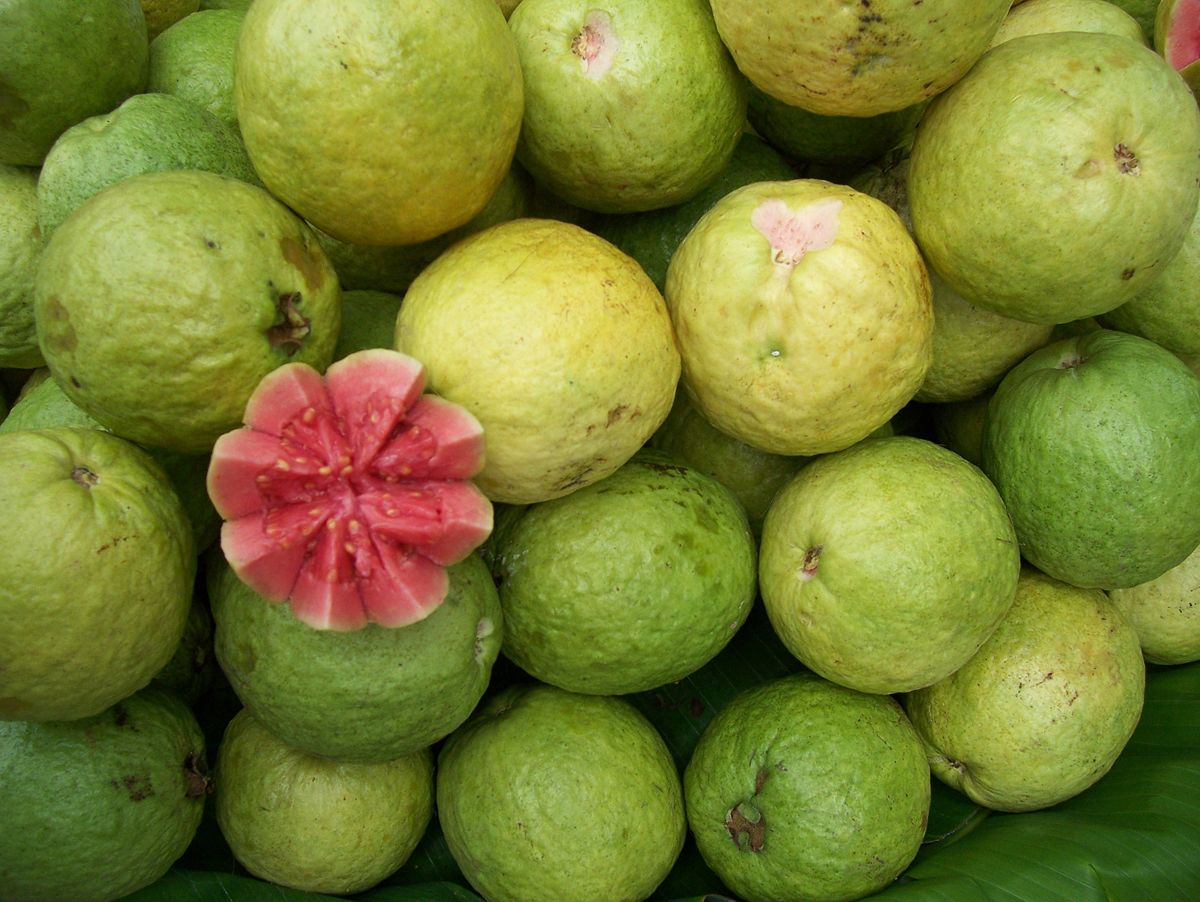}
\caption{Query Image}
\end{subfigure}
\hfill
\begin{subfigure}[b]{0.22\textwidth}
\includegraphics[width=\textwidth]{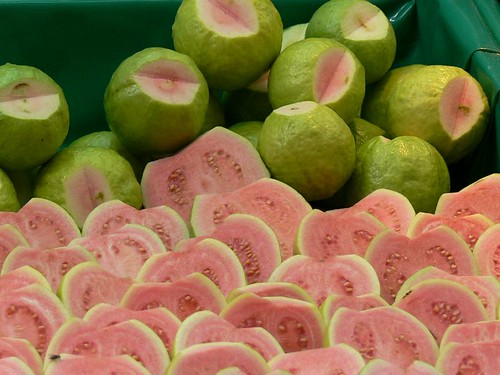}
\caption{mSigLIP (0.923)}
\end{subfigure}
\hfill
\begin{subfigure}[b]{0.22\textwidth}
\includegraphics[width=\textwidth]{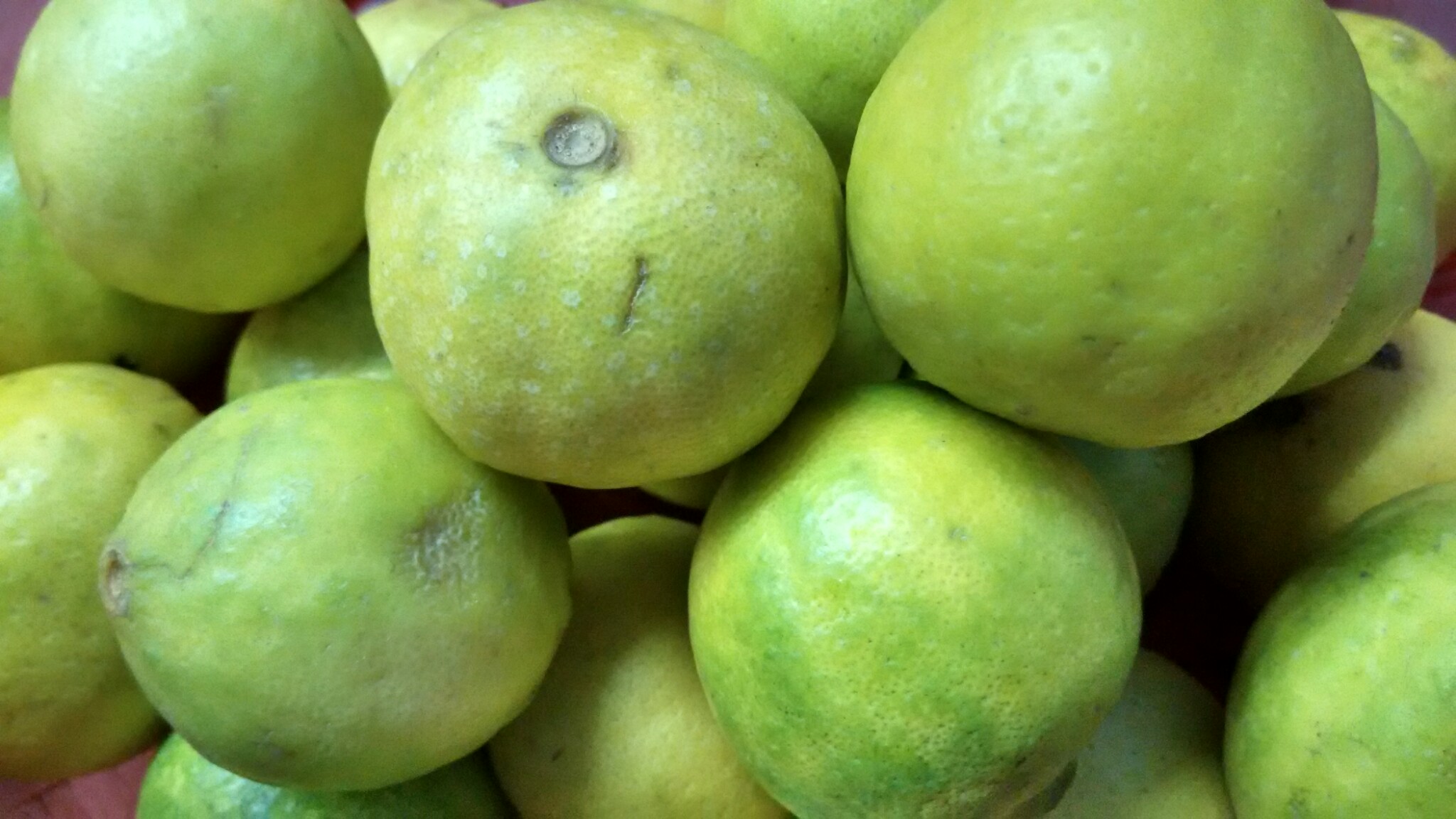}
\caption{CLIP (0.918)}
\end{subfigure}
\hfill
\begin{subfigure}[b]{0.22\textwidth}
\includegraphics[width=\textwidth]{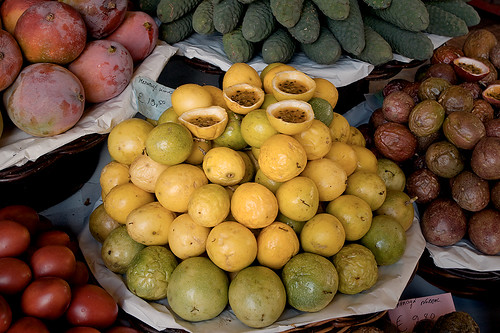}
\caption{DINOv2 (0.850)}
\end{subfigure}
\caption{Retrieval Comparison Across Encoders}
\label{fig:encoder_comp}
\end{figure}

We evaluated several image-text encoder models to compute both image-image and image-text similarity. After qualitatively analyzing the outputs from different variants, we found that mSigLIP performed the best. Its unique loss function improves image-text alignment, making it superior to other encoders such as CLIP \cite{radford2021learning} and DINOv2 \cite{oquab2024dinov2learningrobustvisual} for our task. We conduct a qualitative analysis of retrieval performance using a small subset of images. This subset comprises all fruit images from BabelNet, with each encoder provided the same set of potential matches for retrieval. \autoref{fig:encoder_comp} presents a comparative evaluation of retrieval performance across different encoders, demonstrating mSigLIP’s superior semantic image understanding and similarity scores.

\subsection{Visual Entity Linking}
\label{sec:Vel_appendix}

\begin{table*}[!htp]
\centering
\resizebox{\textwidth}{!}{%
\begin{tabular}{lrrrr rrrr rrrr rrrr}
\toprule
Datasets & \multicolumn{4}{c}{IN-Food} & \multicolumn{4}{c}{IN-Plant} & \multicolumn{4}{c}{IN-Animal} & \multicolumn{4}{c}{IN-Artifact} \\
\cmidrule(lr){2-5} \cmidrule(lr){6-9} \cmidrule(lr){10-13} \cmidrule(lr){14-17}
Methods & @1 & @5 & @10 & @20 & @1 & @5 & @10 & @20 & @1 & @5 & @10 & @20 & @1 & @5 & @10 & @20 \\
\midrule
\texttt{Idefics-2B}        & 11.58 & 10.95 & 10.72 & 11.91 &  4.60 &  4.20 &  4.68 &  5.03 &  8.51 &  6.86 &  8.30 &  9.10 &  7.48 &  8.19 &  8.82 &  9.43  \\
\texttt{Lemma (V-T)}       & \textbf{23.65} & \textbf{50.65} & \textbf{57.93} & \textbf{62.07} & 16.15 & 42.14 & 51.30 & \textbf{57.34} & 22.47 & 52.45 & 62.64 & \textbf{70.11} & \textbf{11.61} & \textbf{22.02} & \textbf{25.51} & \textbf{27.88} \\
\texttt{Freq-Based (V-T)}  & 22.71 & 45.79 & 52.02 & 57.85 & \textbf{26.32} & \textbf{46.21} & \textbf{53.24} & 56.24 & \textbf{31.84} & \textbf{56.55} & \textbf{63.43} & 68.52 & 10.34 & 19.99 & 23.16 & 25.98  \\
\texttt{Lemma (T)}         &  4.68 & 13.69 & 18.02 & 22.78 &  3.20 &  9.13 & 12.82 & 17.10 &  5.67 & 15.66 & 21.69 & 28.17 &  1.63 &  4.15 &  5.65 &  7.22   \\
\texttt{Gloss (T)}         &  3.32 &  8.13 & 11.54 & 15.16 &  2.39 &  7.32 & 10.82 & 15.17 &  4.15 & 12.46 & 17.66 & 23.73 &  1.09 &  2.85 &  3.80 &  4.97   \\
\texttt{Wikipedia (T)}     &  8.40 & 20.48 & 25.93 & 31.83 &  4.99 & 14.52 & 20.50 & 27.83 &  6.84 & 20.76 & 28.60 & 37.20 &  4.38 & 10.72 & 13.84 & 17.03  \\
\bottomrule
\end{tabular}%
}
\caption{Retrieval accuracy at different thresholds across the ImageNet datasets for various disambiguation methods. The highest accuracy for each dataset under each setting is highlighted.}
\label{tab:foci_open}
\end{table*}

In this section, we detail further approaches to perform visual entity linking.

\emph{Image Query + Text Keys}: We compute the similarity between input image embeddings and KB text embeddings to retrieve the top-20 entities, ranking them by similarity score. We build three different text-embedding indices to individually search over for top-matching keys, given the query image. The first comprises of lemma embeddings of each entity, the second comprises of gloss embeddings of each entity,\footnote{\url{https://www.babelnet.org/synset?id=bn:00015267n&orig=dog&lang=EN}} and the third comprises of Wikipedia page embeddings.\footnote{\url{https://www.tensorflow.org/datasets/community_catalog/huggingface/wikipedia}}

\emph{Image Query + Image Keys}: We experiment with bypassing text-based disambiguation and rank each KB ID based on its occurrence frequency among the top-20 retrieved images, a method we refer to as frequency-based matching. 

All of our methods are summarized below:
\begin{enumerate}[noitemsep]
\item \texttt{Lemma (T):} Image-to-text similarity with KB lemmas.
\item \texttt{Gloss (T):} Image-to-text similarity with KB glosses.
\item \texttt{Wikipedia (T):} Image-to-text similarity with Wikipedia content.
\item \texttt{Lemma (V-T):} Vision-based similarity followed by text matching.
\item \texttt{Frequency-Based (V-T):} Most frequently occurring IDs among retrieved images.
\end{enumerate}


To identify the best-performing method among the above alternatives, we conduct a quantitative evaluation using the Fine-grained Object Classification (FOCI) benchmark \cite{geigle2024africaneuropeanswallowbenchmarking}. The FOCI benchmark includes nine datasets, with four subsets of ImageNet (IN-Food, IN-Plant, IN-Animal, IN-Artifact). We select this subset from the FOCI benchmark as we are able to map each ImageNet ID to the corresponding BabelNet ID using the structure of ImageNet-21k. This is enabled by ImageNet being built using WordNet categories and hierarchy. Dataset statistics are listed in \autoref{tab:foci_in_stats}. Each category has 10 associated images.

\begin{table}[h]
    \centering
    \begin{tabular}{|c|c|}
        \hline
        \textbf{\texttt{Dataset}} & \textbf{\texttt{No. of Categories}} \\
        \hline
        \texttt{IN-Food} & 563 \\
        \texttt{IN-Plant} & 957 \\
        \texttt{IN-Animal} & 1314 \\
        \texttt{IN-Artifact} & 2630 \\
        \hline
    \end{tabular}
    \caption{ImageNet dataset Statistics}
    \label{tab:foci_in_stats}
\end{table}

FOCI is a rigorous benchmark that reframes image classification as a series of multiple-choice questions (MCQs). Leveraging CLIP-based similarity, it constructs challenging answer choices by selecting visually and semantically similar instances that are frequently confused with one another. We conduct our evaluation in an open-ended setting, where the model is provided only the input image, without predefined answer choices, making the task significantly more challenging.

Our evaluation aims to:
\begin{itemize}[noitemsep]
    \item Benchmark our system against state-of-the-art VLMs for VEL.
    \item Compare the effectiveness of various retrieval and disambiguation techniques.
    \item Identify retrieval performance thresholds beyond which VEL performance saturates (e.g., Recall@X).
\end{itemize}

To enable quantitative evaluation, we map each category in the IN datasets to BabelNet entities and consider a prediction correct if it retrieves the exact BabelNet ID. This facilitates a direct comparison of different disambiguation strategies. Additionally, by evaluating retrieval at varying levels, we can determine the point at which performance saturates and establish the optimal number of retrieved entities per example.

Results on the FOCI benchmark shown in \autoref{tab:foci_open} in an open setting reveal that image-to-image retrieval combined with image-text disambiguation (\texttt{Lemma (V-T); Gloss (V-T); Frequency-based (V-T)}) consistently surpasses direct image-text retrieval methods (\texttt{Lemma (T); Gloss (T); Wikipedia (T)}). We outperform all state-of-the-art VLMs presented as baselines in the FOCI work itself, demonstrating the robustness of our framework in generally identifying objects belonging to the long tail and linking them to appropriate KB entities.









\subsection{VEL Example}
\label{sec:vel_example}

\begin{figure}[h]
    \centering
    \includegraphics[width=0.8\columnwidth]{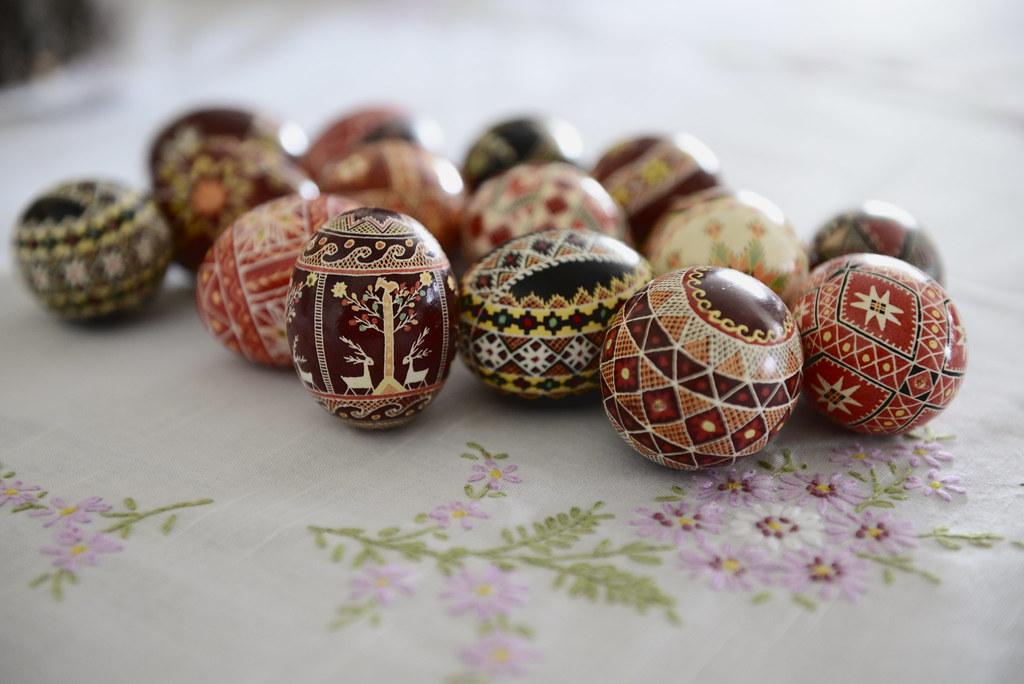}
    \captionsetup{justification=centering}
    \caption{VEL Query Image: \\ Pysanka (Slavic Decorated Egg)}
    \label{fig:VEL_Query_Image}
\end{figure}

\begin{figure*}[h]
\centering
\foreach \col/\score in {1/0.9232446, 2/0.92118037, 3/0.92117214, 4/0.9197761, 5/0.9105648} {
    \begin{minipage}[t]{0.19\textwidth}
    \centering
    \includegraphics[width=\textwidth]{sections/figures/pysanka/pysanka_\col.jpg}
    {\footnotesize Sim: \score}
    \end{minipage}%
}
\caption{Top-5 Image Retrieval Results for Query Image.}
\label{fig:top_5_retrieval}
\end{figure*}

This section aims to provide a detailed example of the best performing entity linking techniques, illustrating various disambiguation techniques along with their respective variations. 

We begin by retrieving the top 20 images from our knowledge base (KB) based on the query image shown in \autoref{fig:VEL_Query_Image}. \autoref{fig:top_5_retrieval} illustrates the 5 highest similarity matches.

\noindent \textbf{Note:} While we present step-by-step results using only 5 retrieved images for clarity, our actual framework utilizes 20 images.

\begin{table}[h]
    \centering
    \begin{tabular}{|c|c|}
        \hline
        \textbf{\texttt{BabelNet ID}} & \textbf{\texttt{Lemma}} \\
        \hline
        \texttt{bn:00068196n}  & \texttt{Romania} \\
        \texttt{bn:00029497n}  & \texttt{Easter} \\
        \texttt{bn:00078872n}  & \texttt{Ukraine} \\
        \texttt{bn:00538675n}  & \texttt{Folklore of Romania} \\
        \texttt{bn:02889635n}  & \texttt{Etymology of Ukraine} \\
        \texttt{bn:03096581n}  & \texttt{Pysanka} \\
        \texttt{bn:00029503n}  & \texttt{Easter egg} \\
        \hline
    \end{tabular}
    \caption{Unique BabelNet IDs and their lemmas.}
    \label{tab:unique_ids}
\end{table}

Each retrieved image in \autoref{fig:top_5_retrieval} corresponds to a list of BabelNet IDs as shown in \autoref{tab:unique_ids}. We extract and aggregate these IDs, retaining only the unique ones before proceeding to the disambiguation stage.  
The unique IDs and their corresponding lemmas are presented below:

After aggregation, these IDs are treated uniformly, irrespective of their initial image similarity.
Our approach incorporates three primary disambiguation techniques.

\noindent \textbf{\texttt{lemma matching:}} We compute the cosine similarity between the query image embeddings and the lemma text embeddings for each BabelNet ID in the retrieved set. The IDs are ranked based on their similarity scores. \autoref{tab:lemma_match} presents the top-5 ranked IDs alongside their corresponding lemmas and similarity scores. As observed, the highest-ranked lemma aligns with the gold label of the query image.

\begin{table}[h]
    \centering
    \resizebox{\columnwidth}{!}{
        \begin{tabular}{|c|c|c|}
            \hline
            \textbf{\texttt{BabelNet ID}} & \textbf{\texttt{Lemma}} & \textbf{\texttt{Score}} \\
            \hline
            \texttt{bn:03096581n}  & \texttt{Pysanka} & \texttt{0.5243} \\
            \texttt{bn:00029497n}  & \texttt{Easter} & \texttt{0.5161} \\
            \texttt{bn:00029503n}  & \texttt{Easter egg} & \texttt{0.5157} \\
            \texttt{bn:00538675n}  & \texttt{Folklore of Romania} & \texttt{0.5108} \\
            \texttt{bn:00068196n}  & \texttt{Romania} & \texttt{0.5037} \\
            \texttt{bn:00078872n}  & \texttt{Ukraine} & \texttt{0.4985} \\
            \texttt{bn:02889635n}  & \texttt{Etymology of Ukraine} & \texttt{0.4942} \\
            \hline
        \end{tabular}
    }
    \caption{Ranked BabelNet IDs}
    \label{tab:lemma_match}
\end{table}

\begin{table}[h]
    \centering
    \resizebox{\columnwidth}{!}{
        \begin{tabular}{|c|c|c|}
            \hline
            \textbf{\texttt{BabelNet ID}} & \textbf{\texttt{Lemma}} & \textbf{\texttt{Frequency}} \\
            \hline
            \texttt{bn:03096581n}  & \texttt{Pysanka} & \texttt{13} \\
            \texttt{bn:00029503n}  & \texttt{Easter egg} & \texttt{10} \\
            \texttt{bn:00029497n}  & \texttt{Easter} & \texttt{6} \\
            \texttt{bn:00078872n}  & \texttt{Ukraine} & \texttt{2} \\
            \texttt{bn:00068196n}  & \texttt{Romania} & \texttt{1} \\
            \texttt{bn:00538675n}  & \texttt{Folklore of Romania} & \texttt{1} \\
            \texttt{bn:02889635n}  & \texttt{Etymology of Ukraine} & \texttt{1} \\
            \hline
        \end{tabular}
    }
    \caption{Frequency ranked BabelNet IDs}
    \label{tab:frequency_match}
\end{table}

\noindent \textbf{\texttt{gloss matching:}} In this method, we utilize the gloss associated with each BabelNet ID. The gloss is a short definition or description of the entity or concept. This method follows the same approach as lemma matching, where we rank IDs according to gloss-image similarity. In this particular case, this method yields a ranking identical to 'lemma matching'.


\noindent \textbf{\texttt{frequency matching:}} This method ranks BabelNet IDs based on their frequency of occurrence within the ID lists associated with the retrieved images. The results of this approach are presented in  \autoref{tab:frequency_match}, which reports the frequencies computed over a set of 20 retrieved images.



\section{Further \method design considerations}

\subsection{Alternative Cultural relevance scoring \mbox{techniques}}\label{sec:alternative_design}

\paragraph{Log-likelihoods:}
In this approach, we opt to directly use the LM's token log-likelihoods corresponding to each candidate culture label.
Given an input prompt consisting of the retrieved textual information about an entity in the image, we prompt the LM with a completion in the form of: ``\textit{This text is relevant to [culture label]}''. 
We estimate the relevance of each culture label directly using the LM head log-likelihood of the completion. 

We use the notation $\mathcal{L}_m(y,x)$ to denote the negative log likelihood of tokens $y$ from model $m$ conditioned on context $X$, or
\begin{equation}
\mathcal{L}_m(y, X) = - \log P_m(y \mid X)
\end{equation}

To compute the log-likelihood of a culture label $c$, we construct context $X$ containing \textit{retrieved} documents $D$, the aforementioned ``\textit{This text is relevant to}'' prompt $p$, and input image $I$ to compute

\begin{equation}
\method (I,c) = \mathcal{L}_m(y, (D_i,I,p))
\end{equation}


One problem with this approach is that models $m$ predict different base rates $\mathcal{L}_m(c_i, (\emptyset,\emptyset,p))$ for different culture symbols when conditioned on the prompt alone.
This introduces a bias where, for example, models will systematically prefer ``\textit{relevant to Australia}'' over ``\textit{relevant to Suriname}''  hindering consistent comparison of attribution scores between cultures. 


To mitigate this, we apply an affine debiasing method using base rates and hyperparameters \(\lambda\) and \(T\) \cite{zhao2021calibrate}. We adjust the likelihood scores by subtracting a scaled correction term while capping its contribution at a threshold to prevent excessive influence from low-likelihood completions, ensuring a balanced adjustment across all outputs:


\small
\begin{equation}
    \hat{\mathcal{L}}(c_i, (D,I,p)) = \mathcal{L}(c_i, (D,I,p)) - \lambda \cdot \max(\mathcal{L}(c_i, (\emptyset,p)), T)
\end{equation}
\normalsize


In practice, we find that ordinal scoring works better than converting raw log-likelihoods to a relevance rating. 1-5 scoring is also practically feasible to correlate with human judgment. The final \method formulation uses ordinal scoring to grade images across cultures in the label set.

\begin{table}[h!]
    \centering
    \setlength{\tabcolsep}{3.7pt} 
    \resizebox{\columnwidth}{!}{
    \begin{tabular}{l cccc c}
        \toprule
        \textbf{Model} & \textbf{Img} & \textbf{Wiki} & \textbf{Top-20} & \textbf{F1} & \textbf{\(\Delta\)$_{\method}$} \\
        \midrule
        \multicolumn{6}{c}{\textbf{\method}} \\
        \midrule
        \multirow{2}{*}{Llama-3.2-11B-Vis.-Ins.} & \checkmark  & \checkmark &  & 47.4 & {\scriptsize (+6.6)} \\ 
        & \checkmark  &  & \checkmark & 47.9 & {\scriptsize (+7.1)} \\
        \hdashline
        \multirow{2}{*}{Qwen2.5-VL-7B-Ins.} & \checkmark  & \checkmark &  & 65.5 & {\scriptsize (+24.5)} \\ 
        & \checkmark  &  & \checkmark & 54.8 & {\scriptsize (+13.8)} \\
        \hdashline
        \multirow{2}{*}{Pangea-7B-hf} & \checkmark  & \checkmark &  & 42.9 & {\scriptsize (+22.6)} \\ 
        & \checkmark  &  & \checkmark & 29.0 & {\scriptsize (+8.7)} \\
        \midrule
        \multicolumn{6}{c}{\textbf{\method (Log-probabilities)}} \\
        \midrule
        Llama-3.2-11B-Vis-Ins. & \checkmark  & \checkmark &  & \textbf{61.8} & {\scriptsize (+21.0)} \\
        Qwen2.5-VL-7B-Ins. & \checkmark  & \checkmark &  & 43.6 & {\scriptsize (+2.6)} \\
        Pangea-7B-hf & \checkmark  & \checkmark &  & 56.0 & {\scriptsize (+35.7)} \\
        \bottomrule
    \end{tabular}
    }
    \caption{
    F1-scores on the \texttt{specific} set. $\Delta_{\method}$ represents the improvement each \method implementation achieves over the strongest baseline. 
    }
    \label{tab:f1alternative}
\end{table}

\section{Experimental Details}

\subsection{Performance of VEL on the \texttt{specific} set}
\label{sec:upper_bound}

In this section, we evaluate the effectiveness of \method in visual entity linking i.e., matching images to the right entities.


To estimate an upper bound on how much VEL performance impacts the overall F1 score in the case of the \texttt{specific set}, we incorporate gold-standard Wikipedia pages as context in our cultural relevance scoring, allowing us to quantify retrieval-induced loss.

\begin{table}[t]
    \centering
    \scriptsize
    \renewcommand{\arraystretch}{1.2}
    \setlength{\tabcolsep}{10pt}
    \resizebox{\linewidth}{!}{%
    \begin{tabular}{l c r}
        \toprule
        \textbf{Model} & \textbf{Image} & \textbf{Ratio} \\
        \midrule
        \multicolumn{3}{c}{\textbf{\method (1-5 Scores)}} \\
        \midrule
        Llama-3.2-11B-Vision-Instruct & \checkmark & 97.1 \\
        Qwen2.5-VL-7B-Instruct & \checkmark & 84.3 \\
        Pangea-7B-hf & \checkmark & 92.1 \\
        \midrule
        \multicolumn{3}{c}{\textbf{\method (Log-probabilities)}} \\
        \midrule
        Llama-3.2-11B-Vision-Instruct & \checkmark & 94.1 \\
        Qwen2.5-VL-7B-Instruct & \checkmark & 96.7 \\
        Pangea-7B-hf & \checkmark & 89.5 \\
        \bottomrule
    \end{tabular}
    }
    \caption{Ratio between \method \space and Gold context}
    \label{tab:gold_ratios}
\end{table}

We obtain the results in \autoref{tab:gold_results_synset} by replacing the retrieved Wikipedia pages with the ground-truth pages corresponding to the entities in our \texttt{specific set}. Specifically, we compute the ratio of the F1-scores from our VEL system to those obtained using the gold-standard pages. As shown in \autoref{tab:gold_ratios}, our method consistently achieves scores comparable to the gold standard, highlighting the effectiveness of our retrieval system in identifying culturally relevant entities.

\begin{table}[t]
    \centering
    \setlength{\tabcolsep}{3.7pt} 
    \resizebox{\columnwidth}{!}{%
    \begin{tabular}{l cc r} 
        \toprule
        \textbf{Model} & \textbf{Image} & \textbf{Wikipedia} & \textbf{F1 Score} \\
        \midrule
        \multicolumn{4}{c}{\textbf{\method (1-5 Scores)}} \\
        \midrule
        \multirow{1}{*}{Llama-3.2-11B-Vision-Instruct} & \checkmark  & \checkmark & 48.8 \\
        \multirow{1}{*}{Qwen2.5-VL-7B-Instruct} & \checkmark  & \checkmark & \textbf{77.7} \\
        \multirow{1}{*}{Pangea-7B-hf} & \checkmark  & \checkmark & 46.6 \\
        \midrule
        \multicolumn{4}{c}{\textbf{\method (Log-probabilities)}} \\
        \midrule
        Llama-3.2-11B-Vision & \checkmark  & \checkmark & \textbf{65.7} \\
        Qwen2.5-VL-7B-Instruct & \checkmark  & \checkmark & 45.1 \\
        Pangea-7B-hf & \checkmark  & \checkmark & 62.6 \\
        \bottomrule
    \end{tabular}}
    \caption{F1-scores with Gold context}
    \label{tab:gold_results_synset}
\end{table}

\subsection{Culturally Similar Matches}
\label{sec:cul_sim_appendix}

\begin{figure}[htbp]
    \renewcommand{\arraystretch}{1.5} 
    \centering
    \begin{tabular}{p{0.38\columnwidth} p{0.5\columnwidth}} 
        \includegraphics[width=0.38\columnwidth]{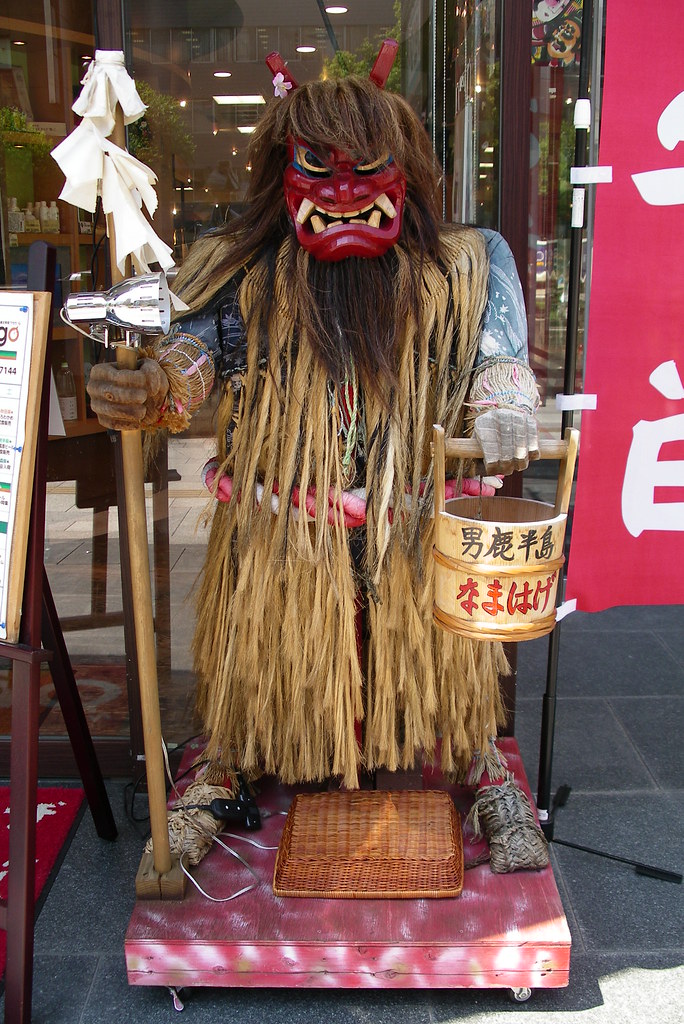} &
        \raisebox{0.5\height}{\parbox{0.5\columnwidth}{\centering \textbf{Gold:} Namahage \\ \textbf{\method:} Oni}} \\[10pt]

        \includegraphics[width=0.38\columnwidth]{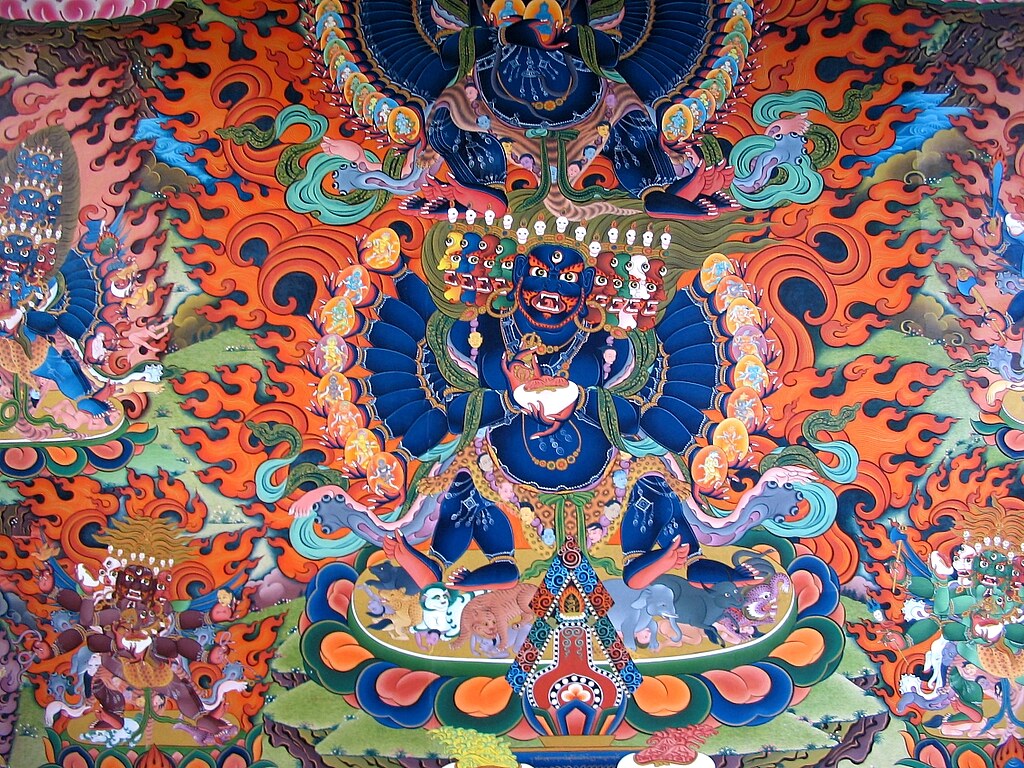} &
        \raisebox{0.5\height}{\parbox{0.5\columnwidth}{\centering \textbf{Gold:} Thangka Painting \\ \textbf{\method:} Tibetan Art}} \\[10pt]

        \includegraphics[width=0.38\columnwidth]{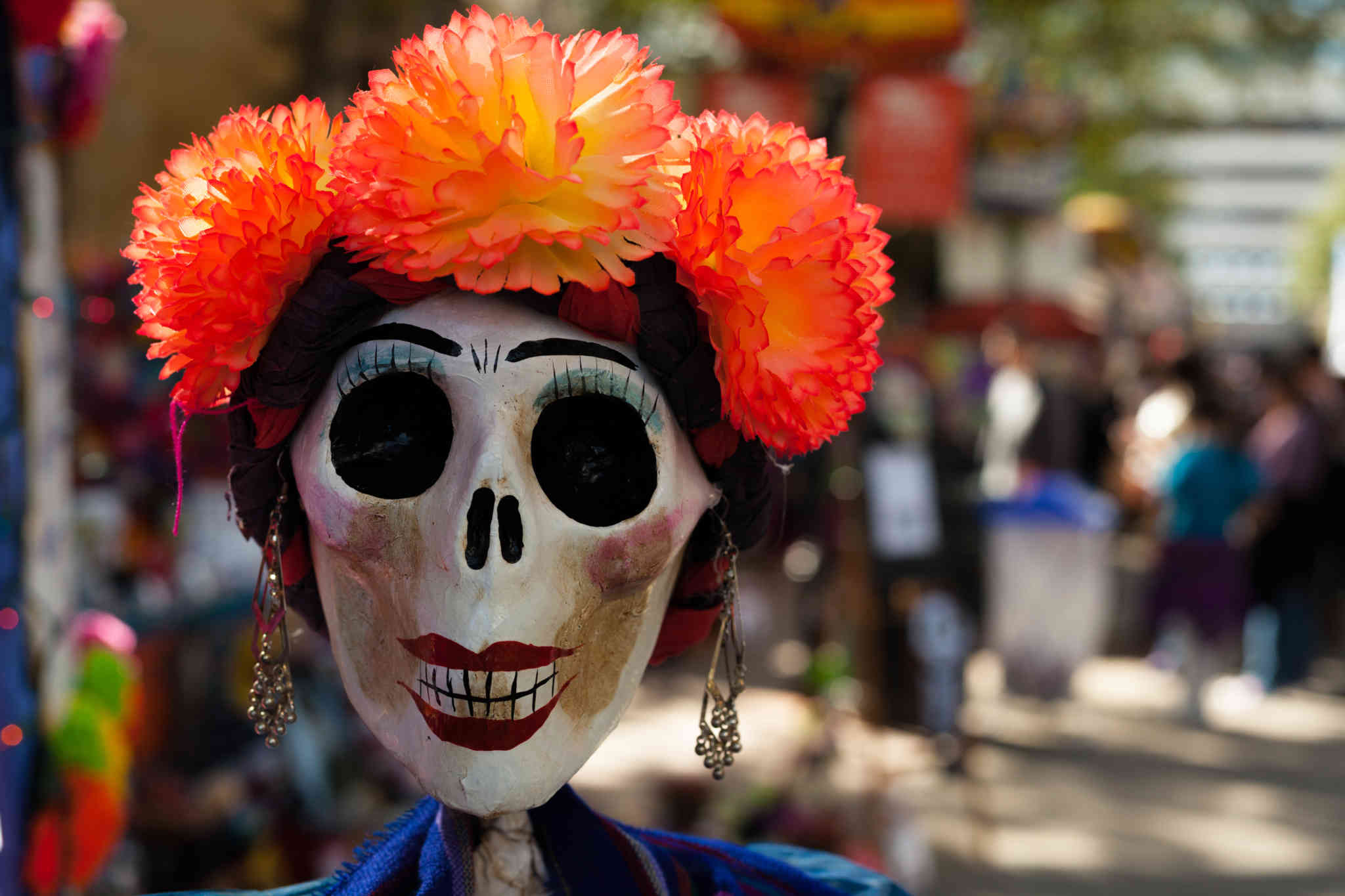} &
        \raisebox{0.5\height}{\parbox{0.5\columnwidth}{\centering \textbf{Gold:} Day of the Dead \\ \textbf{\method:} Mictēcacihuātl}} \\
    \end{tabular}
    \caption{Specific set examples}
    \label{fig:cul_sim_matches}
\end{figure}

There are instances where the exact entity is not matched. A few of these examples are presented in \autoref{fig:cul_sim_matches}.

\noindent \textbf{Case 1:} In this case, the retrieved entity represents a specific aspect of a broader cultural concept. For example, as illustrated in the first case, the retrieved entity is an Oni, a type of Japanese troll, whereas the gold-standard label, Namahage, refers to a broader category of Japanese demon-like creatures. Despite this distinction, the retrieved entity remains culturally relevant and semantically aligned with the gold label.  

\noindent \textbf{Case 2:} The second case exemplifies instances where the retrieved entity directly represents the associated culture. The image depicts a Thangka painting, a form of Tibetan art. Our retrieval system identifies the broader concept of Tibetan art, which remains an accurate cultural attribution given the overarching goal of cultural relevance assessment.  

\noindent \textbf{Case 3:} The final case mirrors the first, where the retrieved entity corresponds directly to the deity depicted in the image. This deity is highly relevant and closely associated with the gold-standard label, which represents the broader cultural event, the Day of the Dead.  

\noindent In such cases, while the retrieval is technically classified as incorrect, the overall cultural scoring remains reliable. This is because the retrieved entities are often culturally analogous to the target entities, and their corresponding Wikipedia pages contain thematically similar content. Consequently, the models can still effectively extract the relevant cultural information.

\subsection{GlobalRG Concept List}
\label{sec:globalrg_concepts}

\autoref{tab:cultural_categories} presents the 20 human universals selected by \cite{bhatia2024local}.

\begin{table}[htbp]
    \centering
    \renewcommand{\arraystretch}{1.2}
    \resizebox{\columnwidth}{!}{
    \begin{tabular}{|l |l |l |l|}
        \hline
        Breakfast & Clothing & Dance & Drinks \\
        \hline
        Dessert & Dinner & Farming & Festival \\
        \hline
        Eating Habits & Funeral & Greetings & Head Coverings \\
        \hline
        Instrument & Lunch & Marriage & Music \\
        \hline
        Religion & Ritual & Sports & Transport \\
        \hline
    \end{tabular}
    }
    \caption{Cultural Categories Table}
    \label{tab:cultural_categories}
\end{table}

\subsection{Cultural Relevance Scoring Criteria}
\label{sec:Relevance_Score_appendix_section} \autoref{tab:Cultural_relevance_scores} presents descriptions for each score in our 1-5 scoring formulation.

\begin{table}[htbp]
\centering
\renewcommand{\arraystretch}{1.2}
\setlength{\tabcolsep}{4pt}
\resizebox{\linewidth}{!}{%
\begin{tabular}{c c p{6.5cm}}
\toprule
\textbf{Score} & \textbf{Relevance Level} & \textbf{Description} \\
\midrule
\textbf{1} & Not Relevant & The content does not connect with or reflect the target culture at all. \\
\addlinespace[2pt]
\textbf{2} & Minimally Relevant & The content shows slight or superficial connections to the culture but lacks depth. May include vague references or isolated cultural elements that feel out of place or underdeveloped. \\
\addlinespace[2pt]
\textbf{3} & Somewhat Relevant & The content contains identifiable cultural references, but they may feel generic, inconsistent, or limited in scope. The connection to the culture is present but could be stronger or more meaningful. \\
\addlinespace[2pt]
\textbf{4} & Relevant & The content reflects a reasonable understanding of the culture, including accurate and appropriate references. It integrates cultural aspects well, though there may still be areas where more depth could be added. \\
\addlinespace[2pt]
\textbf{5} & Highly Relevant & The content is deeply connected to the target culture, showing an immersive, accurate, and respectful understanding. Cultural references feel natural, meaningful, and central to the content. \\
\bottomrule
\end{tabular}
}
\caption{Cultural Relevance Scoring Criteria}
\label{tab:Cultural_relevance_scores}
\end{table}


\subsection{Example Prompt for 1-5 Scoring}
A prompt template used in the 1-5 Scoring method for VLMs is provided in \autoref{fig:template_prompt}.

\definecolor{entityColor}{RGB}{0, 102, 204}   
\definecolor{wikiColor}{RGB}{204, 0, 102}     
\definecolor{targetColor}{RGB}{0, 153, 51} 

\begin{figure*}[h]
    \centering
    \begin{tcolorbox}[colframe=black!70, colback=gray!10, sharp corners]
        \textbf{Prompt Template:} \\
        We want to assess how relevant an image is to a given culture.\\
        We have identified this concept to be closely associated with the image: 
        \textcolor{entityColor}{\texttt{\{entity\}}}. \\
        Here is some detailed information about this concept from Wikipedia: 
        \textcolor{wikiColor}{\texttt{\{wiki\}}}.\\
        
        Using the above context, assign a score from 1 to 5 based on how culturally relevant the image is to 
        \textcolor{targetColor}{\texttt{\{target\}}}: \\
        Think step by step, specifically considering cultural symbols, styles, traditions, or any features that align with the culture of 
        \textcolor{targetColor}{\texttt{\{target\}}}.\\

        The final score should be a number between 1 to 5, where the meaning of each score is defined as follows:

        \begin{itemize}[noitemsep]
            \item 1 -- Not Relevant: The content does not connect with or reflect the target culture at all.
            \item 2 -- Minimally Relevant: The content shows slight or superficial connections to the culture but lacks depth. May include vague references or isolated cultural elements that feel out of place or underdeveloped.
            \item 3 -- Somewhat Relevant: The content contains identifiable cultural references, but they may feel generic, inconsistent, or limited in scope. The connection to the culture is present but could be stronger or more meaningful.
            \item 4 -- Relevant: The content reflects a reasonable understanding of the culture, including accurate and appropriate references. It integrates cultural aspects well, though there may still be areas where more depth could be added.
            \item 5 -- Highly Relevant: The content is deeply connected to the target culture, showing an immersive, accurate, and respectful understanding. Cultural references feel natural, meaningful, and central to the content.
        \end{itemize}

        The output should be a single number ONLY.
    \end{tcolorbox}

    \vspace{0.2em} 

    \begin{tcolorbox}[colframe=entityColor, colback=gray!10, sharp corners]
        \textbf{Example Conversation Format:} \\

        \textbf{System:} \textit{You are an expert in evaluating the cultural relevance of images.} \\

        \textbf{User:}  
        \texttt{[Image]}  
        \texttt{[Text: Prompt Template]}  

        \textbf{Model:} \\   
        \textbf{Final Score:} \texttt{[1-5]}
    \end{tcolorbox}

\caption{Prompt Template for \method \space methods}
\label{fig:template_prompt}
\end{figure*}

\subsection{Specific Set Label Set}
\label{sec:specific_set_labels}

\autoref{tab:spec_cultural_labels} shows the full set of candidate culture labels for the images in the \texttt{Specific Set:}

\begin{table}[h]
    \centering
    \tiny
    \resizebox{\columnwidth}{!}{%
    \begin{tabular}{ll}
        \toprule[0.4pt]
        \textbf{Category} & \textbf{Labels} \\
        \midrule[0.3pt]
        \multirow{9}{*}{\textbf{Geographical Entities}} & Countries of the world \\
        & Countries of Africa \\
        & States of Mexico \\
        & States of the US \\
        & States of India \\
        & Major cities of India \\
        & Cities of Indonesia \\
        & Cities of Nepal \\
        \midrule[0.3pt]
        \multirow{5}{*}{\textbf{Ethnic \& Cultural Groups}} & Ethnicities of Africa \\
        & Ethnicities of India \\
        & Ethnicities of Australia \\
        & Ethnicities of the US \\
        & Ethnicities of Indonesia \\
        \midrule[0.3pt]
        \textbf{Festivals \& Traditions} & Festivals of India \\ 
        \midrule[0.3pt]
        \multirow{2}{*}{\textbf{Philosophy \& Religion}} & Japanese philosophy \\
        & World religions \\
        \midrule[0.3pt]
        \textbf{Historical Civilizations} & Bronze Age civilizations \\ 
        \bottomrule[0.4pt]
    \end{tabular}%
    }
    \caption{Cultural labels for the Specific Set}
    \label{tab:spec_cultural_labels}
\end{table}

\subsection{Alternate Thresholds}
\label{sec:alt_thresholds}

\autoref{tab:thresholds} provides the F1-score, Precision and Recall at different score thresholds for binary classification on the \texttt{specific} set. We note that drops in F1-score are primarily driven by drops in Precision. 

\begin{table}[h]
\centering
\resizebox{\linewidth}{!}{
\begin{tabular}{llccc ccc ccc ccc}
\toprule
& \textbf{Method} 
& \multicolumn{3}{c}{\textbf{Threshold=2}} 
& \multicolumn{3}{c}{\textbf{Threshold=3}} 
& \multicolumn{3}{c}{\textbf{Threshold=4}} 
& \multicolumn{3}{c}{\textbf{Threshold=5}} \\
\cmidrule(lr){3-5} \cmidrule(lr){6-8} \cmidrule(lr){9-11} \cmidrule(lr){12-14}
& & \textbf{F1} & \textbf{P} & \textbf{R} 
  & \textbf{F1} & \textbf{P} & \textbf{R} 
  & \textbf{F1} & \textbf{P} & \textbf{R} 
  & \textbf{F1} & \textbf{P} & \textbf{R} \\
\midrule
\multicolumn{14}{l}{\textbf{CAIRE}} \\
& Qwen-LM   & 0.33 & 0.17 & 0.96 & 0.40 & 0.33 & 0.88 & 0.69 & 0.74 & 0.76 & 0.59 & 0.73 & 0.55 \\
& Qwen      & 0.52 & 0.47 & 0.84 & 0.63 & 0.61 & 0.79 & 0.66 & 0.74 & 0.65 & 0.20 & 0.25 & 0.18 \\
& Pangea    & 0.38 & 0.29 & 0.91 & 0.38 & 0.29 & 0.91 & 0.43 & 0.34 & 0.90 & 0.32 & 0.40 & 0.32 \\
& LLaMA     & 0.34 & 0.31 & 0.82 & 0.48 & 0.47 & 0.76 & 0.47 & 0.51 & 0.60 & 0.00 & 0.00 & 0.00 \\
\addlinespace
\multicolumn{14}{l}{\textbf{Baseline}} \\
& Qwen      & 0.33 & 0.26 & 0.89 & 0.38 & 0.32 & 0.82 & 0.41 & 0.38 & 0.67 & 0.17 & 0.20 & 0.17 \\
& Pangea    & 0.19 & 0.12 & 0.96 & 0.19 & 0.12 & 0.96 & 0.20 & 0.13 & 0.96 & 0.07 & 0.08 & 0.09 \\
& LLaMA     & 0.12 & 0.07 & 1.00 & 0.29 & 0.20 & 0.95 & 0.41 & 0.39 & 0.75 & 0.00 & 0.00 & 0.00 \\
\bottomrule
\end{tabular}
}
\caption{Performance across thresholds for CAIRE and Baseline methods using various VLMs.}
\label{tab:thresholds}
\end{table}

\subsection{Alternate VL-Encoders}
\label{sec:openclip}

We run Baseline-1 using OpenCLIP variants trained on the DFN datasets \cite{fang2023data}. Specifically, the DFN-2B variant achieves a score of 7.0\%\footnote{\url{https://huggingface.co/apple/DFN2B-CLIP-ViT-L-14}}, the DFN-5B variant achieves 8.5\%\footnote{\url{https://huggingface.co/apple/DFN5B-CLIP-ViT-H-14}}, and the DFN-5B-378 variant achieves 8.9\%\footnote{\url{https://huggingface.co/apple/DFN5B-CLIP-ViT-H-14-378}}. These models outperform CLIP ViT-B/32, but achieve lower F1-scores than mSigLIP. Moreover, our approach is inherently modular and can therefore seamlessly integrate with the latest and strongest vision-language encoders as they become available.

\subsection{Human Annotation Details}
\label{sec:hum_ann}

We initially hired 20 annotators per country. After filtering based on annotation quality and completion, the final number of usable annotators varied across countries, as summarized in \autoref{tab:annotators-per-country} for both the \texttt{universal-generated} and \texttt{universal-retrieved} subsets. For each image, we obtain 10 relevance scores from \method \space --- one for each country as the cultural label, and 10 human scores --- computed by averaging responses from annotators of that country to our relevance prompt (based on a 5-point Likert scale). These paired human and model scores are then used to compute the image-level Pearson correlation.

\begin{table}[h]
\centering
\resizebox{\columnwidth}{!}{%
\begin{tabular}{lcc}
\toprule
\textbf{Country}    & \textbf{universal-generated} & \textbf{universal-retrieved} \\
\midrule
Brazil              & 20 & 14 \\
China               & 9  & 15 \\
Egypt               & 9  & 13 \\
Germany             & 9  & 15 \\
India               & 20 & 15 \\
Indonesia           & 9  & 14 \\
Mexico              & 19 & 14 \\
Nigeria             & 17 & 15 \\
Russia              & 10 & 15 \\
United States       & 20 & 15 \\
\bottomrule
\end{tabular}
}
\caption{Number of annotators per country used for the two subsets (after filtering).}
\label{tab:annotators-per-country}
\end{table}

\end{document}